\documentclass{article}
\pdfoutput=1
\PassOptionsToPackage{numbers,sort&compress}{natbib}
\usepackage[final]{neurips_2019}

\usepackage[utf8]{inputenc} % allow utf-8 input
\usepackage[T1]{fontenc}    % use 8-bit T1 fonts
\usepackage{hyperref}       % hyperlinks
\usepackage{url}            % simple URL typesetting
\usepackage{booktabs}       % professional-quality tables
\usepackage{amsfonts}       % blackboard math symbols
\usepackage{nicefrac}       % compact symbols for 1/2, etc.
\usepackage{microtype}      % microtypography
\usepackage{graphicx}
\usepackage[acronym,smallcaps,nowarn,section,nogroupskip,nonumberlist]{glossaries}
\usepackage{amsthm}
\usepackage{amssymb}
\usepackage{color}
\usepackage{todonotes}
\usepackage{enumitem}
\usepackage{caption}
\usepackage{wrapfig}
\usepackage{enumitem}
\usepackage[nameinlink,capitalise]{cleveref}

\glsdisablehyper{}
\makenoidxglossaries
\newacronym{TVO}{tvo}{thermodynamic variational objective}
\newacronym{TVI}{tvi}{thermodynamic variational identity}
\newacronym{AIS}{ais}{annealed importance sampling}
\newacronym{TI}{ti}{thermodynamic integration}
\newacronym{WS}{ws}{wake-sleep}
\newacronym{RWS}{rws}{reweighted wake-sleep}
\newacronym{ELBO}{elbo}{evidence lower bound}
\newacronym{EUBO}{eubo}{evidence upper bound}
\newacronym{VAE}{vae}{variational autoencoder}
\newacronym{IWAE}{iwae}{importance weighted autoencoder}
\newacronym{KL}{kl}{Kullback-Leibler}
\newacronym{SGD}{sgd}{stochastic gradient descent}
\newacronym{VIMCO}{vimco}{variational inference for Monte Carlo objectives}
\newacronym{IS}{is}{importance sampling}
\newacronym{VI}{vi}{variational inference}

\DeclareMathOperator*{\argmin}{arg\,min} % * allows typesetting beneath
\DeclareMathOperator*{\argmax}{arg\,max} % * allows typesetting beneath
\newcommand{\given}{\lvert}
\DeclareMathOperator{\ELBO}{\textsc{elbo}}
\DeclareMathOperator{\TVO}{\textsc{TVO}}

\DeclareMathOperator{\EUBO}{\textsc{eubo}}

\newcommand{\KL}[2]{\textsc{kl}\left(#1 \middle| \middle| #2\right)}
\DeclareMathOperator{\E}{\mathbb{E}}

\DeclareMathOperator{\R}{\mathbb{R}}
\DeclareMathOperator{\Cov}{\mathrm{Cov}}
\DeclareMathOperator{\Var}{\mathrm{Var}}

\DeclareMathOperator{\vx}{\mathbf{x}}
\DeclareMathOperator{\vz}{\mathbf{z}}
\DeclareMathOperator{\vdz}{d\mathbf{z}}
\DeclareMathOperator{\vdx}{d\mathbf{x}}

\Crefname{algocf}{Algorithm}{Algorithms}
\crefname{algorithm}{Algorithm}{Algorithms}
\crefname{equation}{Equation}{Equations}
\crefname{figure}{Figure}{Figure}
\crefname{section}{§}{§§}
\Crefname{section}{§}{§§}
\crefformat{equation}{(#2#1#3)}
\crefformat{section}{§#2#1#3}

\newtheorem{lemma}{Lemma}

\usepackage{multirow}
\usepackage{makecell}

\usepackage[export]{adjustbox}% http://ctan.org/pkg/adjustbox

\title{The Thermodynamic Variational Objective}

\author{
    Vaden Masrani$^1$, Tuan Anh Le$^2$, Frank Wood$^1$\\
    $^1$Department of Computer Science, University of British Columbia\\
    $^2$Department of Brain and Cognitive Sciences, MIT
}

\begin{document}

\maketitle

\begin{abstract}
    We introduce the \gls{TVO} for learning in both continuous and discrete deep generative models. The \gls{TVO} arises from a key connection between variational inference and thermodynamic integration that results in a tighter lower bound to the log marginal likelihood than the standard variational \gls{ELBO} while remaining as broadly applicable. We provide a computationally efficient gradient estimator for the \gls{TVO} that applies to continuous, discrete, and non-reparameterizable distributions and show that the objective functions used in variational inference, variational autoencoders, wake sleep, and inference compilation are all special cases of the \gls{TVO}. We use the \gls{TVO} to learn both discrete and continuous deep generative models and empirically demonstrate state of the art model and inference network learning.
\end{abstract}

\glsresetall

% !TEX root =  tvo.tex

\section{Introduction}
Unsupervised learning in richly structured deep latent variable models~\citep{kingma2014auto,rezende2014stochastic} remains challenging.
Fundamental research directions include low-variance gradient estimation for discrete and continuous latent variable models~\citep{mnih2014neural,mnih2016variational,tucker2017rebar,naesseth2017reparameterization,figurnov2018implicit}, tightening variational bounds in order to obtain better model learning~\citep{burda2016importance,maddison2017filtering,le2018autoencoding,naesseth2017variational}, and alleviation of the associated detrimental effects on learning of inference networks~\citep{rainforth2018tighter}.

We present the \gls{TVO}, which is based on a key connection we establish between \gls{TI} and amortized \gls{VI}, namely that by forming a geometric path between the model and inference network, the ``instantaneous \textsc{elbo}''~\citep{blei_presentation} that appears in \gls{VI} is equivalent to the first derivative of the potential function that appears in \gls{TI}~\citep{gelman1998simulating,neal1993probabilistic}.
This allows us to formulate the log evidence as a 1D integration of the instantaneous \textsc{elbo} in a unit interval, which we then approximate to form the \gls{TVO}.

We demonstrate that optimizing the \gls{TVO} leads to improved learning of both discrete and continuous latent-variable deep generative models.
The gradient estimator we derive for optimizing the \gls{TVO} has empirically lower variance than the \textsc{reinforce}~\citep{williams1992simple} estimator, and unlike the reparameterization trick (which is only applicable to a limited family of continuous latent variables), applies to both continuous and discrete latent variables models.

The \gls{TVO} is a lower bound to the log evidence which can be made arbitrarily tight.
We empirically show that optimizing the \gls{TVO} results in better inference networks than optimizing the \gls{IWAE} objective~\citep{burda2016importance} for which tightening of the bound is known to make inference network learning worse~\citep{rainforth2018tighter}.
While this problem can be ameliorated by reducing the variance of the gradient estimator in the case of reparameterizable latent variables~\citep{tucker2018doubly}, resolving it in the case of non-reparameterizable latent variables currently involves alternating optimization of model and inference networks~\citep{hinton1995wake,bornschein2015reweighted,le2018revisiting}.

\begin{figure}[t!]
    \centering
    \includegraphics[width=\textwidth]{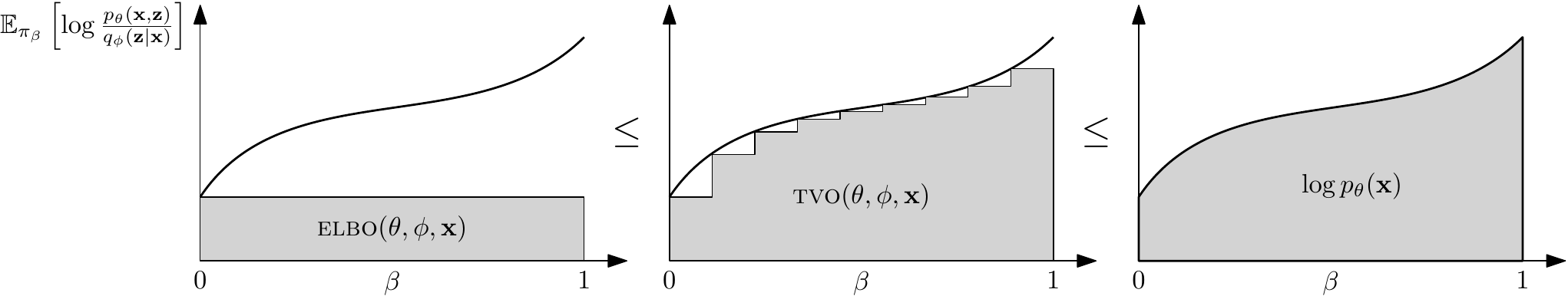}
    \caption{The thermodynamic variational objective (\gls{TVO}) (center) is a finite sum numerical approximation to $\log p_\theta(\vx)$, defined by the \gls{TVI} (right). The $\ELBO$ (left) is a single partition approximation of the same integral. $\pi_\beta$ is given in~\cref{eq:path}}
    \label{fig:introduction/tvo}
  \end{figure}

\section{The Thermodynamic Variational Objective}
\label{sec:method/tvo}
The \gls{ELBO}, used in learning \glspl{VAE}, lower bounds the log evidence of a generative model $p_\theta(\vx, \vz)$ parameterized by $\theta$ of a latent variable $\vz$ and data $\vx$. It can be written as the log evidence minus a \gls{KL} divergence
\begin{align}
    \ELBO(\theta, \phi, \vx) := \log p_\theta(\vx) - \KL{q_\phi(\vz \given \vx)}{p_\theta(\vz \given \vx)}, \label{eq:introduction/elbo}
\end{align}
where $q_\phi(\vz \given \vx)$ is an inference network parameterized by $\phi$.
As illustrated in \cref{fig:introduction/tvo}, the \gls{TVO}
\begin{align}
    {\underbrace{\frac{1}{K}\left[\ELBO(\theta, \phi, \vx) + \sum_{k=1}^{K-1}\E_{\pi_{\beta_{k}}}\left[\log \frac{p_{\theta}(\vx, \vz)}{q_{\phi}(\vz | \vx)}\right]\right]}_{\TVO(\theta, \phi, \vx)}}
    \leq
    {\underbrace{\vphantom{\sum_{k=1}^{K-1}} \int_0^1 \E_{\pi_{\beta}}\left[ \log \frac{p_{\theta}(\vx,\vz)}{q_{\phi}(\vz | \vx)} \right] d\beta = \log p_\theta(\vx)}_{\textsc{thermodynamic variational identity}}}
     \label{eq:tvo_lower}
\end{align}
lower bounds the log evidence by using a Riemann sum approximation to the \gls{TVI}, a one-dimensional integral over a scalar $\beta$ in a unit interval which evaluates to the log model evidence $\log p_\theta(\vx)$.

The integrand, which is a function of $\beta$, is an expectation of the so-called ``instantaneous \gls{ELBO}''~\citep{blei_presentation} under $\pi_\beta(\vz)$, a geometric combination of $p_\theta(\vx, \vz)$ and $q_\phi(\vz \given \vx)$ which we formally define in \cref{sec:method/connection}.
Remarkably, at $\beta = 0$, the integrand equals the \gls{ELBO}. This therefore allows us to view the \gls{ELBO} as a single-term left Riemann sum of the \gls{TVI}.
At $\beta = 1$, the integrand equals to the \gls{EUBO}. This sheds a new unifying perspective on the \gls{VAE} and wake-sleep objectives, which we explore in detail in~\cref{sec:method/generalizations} and \cref{app:special_cases}.

%!tex root=./tvo.tex
\section{Connecting Thermodynamic Integration and Variational Inference}
\label{sec:method/connection}
Suppose there are two unnormalized densities $\tilde \pi_i(\vz)$ ($i = 0, 1$) and corresponding normalizing constants $Z_i := \int \tilde \pi_i(\vz) \vdz$, which together define the normalized densities $\pi_i(\vz) := \tilde \pi_i(\vz) / Z_i$.
We can typically evaluate the unnormalized densities but cannot evaluate the normalizing constants.

While calculating the normalizing constants individually is usually intractable, \glsfirst{TI}~\citep{gelman1998simulating,neal1993probabilistic} allows us to compute the log of the ratio of the normalizing constants, $\log Z_1 / Z_0$. To do so, we first form a family of unnormalized densities (or a ``path'') parameterized by $\beta \in [0, 1]$ between the two distributions of interest
\begin{align}
    \tilde \pi_\beta(\vz) := \tilde{\pi}_1(\vz)^{\beta}\tilde{\pi}_0(\vz)^{{1 - \beta}}
\end{align}
with the corresponding normalizing constants and normalized densities
\begin{align}
  Z_\beta := \int \tilde \pi_\beta(\vz) \vdz,
  \quad
  \text{and}
  \quad
  \pi_\beta(\vz) := \tilde \pi_\beta(\vz) / Z_\beta.
\end{align}
Following~\citet{neal1993probabilistic}, we will find it useful to define a potential energy function ${U}_{\beta}(\vz) := \log \tilde{\pi}_{\beta}(\vz)$ along with its first derivative ${U}'_{\beta}(\vz) = \frac{d{U}_{\beta}(\vz)}{d\beta}$.
We can then estimate the log of the ratio of the normalizing constants via the identity central to \gls{TI}, derived in \cref{app:TI},
\begin{align}
    \log Z_1 - \log Z_0 &= \int_0^1 \E_{\pi_{\beta}}\big[{U}'_{\beta}(\vz)\big] d  \beta. \label{eq:ti_ratio}
\end{align}

Our key insight connecting \gls{TI} and \gls{VI} is the following.
If we set
\begin{equation}
    \begin{aligned}
        &\tilde{\pi}_{0}(\vz):=q_{\phi}(\vz|\vx)
        \quad
        Z_{0}=\int q_{\phi}(\vz | \vx) \vdz =1\\
        &\tilde{\pi}_{1}(\vz):=p_{\theta}(\vx, \vz)
        \quad
        Z_{1}=\int p_{\theta}(\vx,\vz) \vdz =p_{\theta}(\vx)
    \end{aligned}
\end{equation}
this results in a geometric path between the variational distribution $q_\phi(\vz \given \vx)$ and the model $p_\theta(\vx, \vz)$
\begin{align}
  \tilde \pi_\beta(\vz) &:= p_\theta(\vx, \vz)^\beta q_\phi(\vz \given \vx)^{1 - \beta}
  \quad
  \text{and}
  \quad
  \pi_\beta(\vz) := \frac{\tilde \pi_\beta(\vz)}{Z_\beta}, \label{eq:path}
\end{align}
where the first derivative of the potential is equal to the ``instantaneous \gls{ELBO}''~\citep{blei_presentation}
\begin{align}
  U'_\beta(\vz)
  = \log \frac{p_\theta(\vx, \vz)}{q_\phi(\vz \given \vx)}. \label{eq:U_prime}
\end{align}
Substituting \eqref{eq:U_prime} and $Z_0=1$ and $Z_1=p_\theta(\vx)$ into \eqref{eq:ti_ratio} results in the \emph{thermodynamic variational identity}
\begin{align}
  \log p_\theta(\vx) &=  \int_0^1 \E_{\pi_{\beta}}\left[ \log \frac{p_{\theta}(\vx,\vz)}{q_{\phi}(\vz | \vx)} \right] d\beta. \label{eq:tvo/tv_identity}
\end{align}
This means that $\log p_\theta(\vx)$ can be expressed as a one-dimensional integral of an expectation of the instantaneous \gls{ELBO} under $\pi_\beta$ from $\beta = 0$ to $\beta = 1$ (see \cref{fig:introduction/tvo} (right)).

To obtain the \glsreset{TVO}\gls{TVO} defined in \eqref{eq:tvo_lower}, we lower bound the integral in \eqref{eq:tvo/tv_identity} using a left Riemann sum.
That this is in fact a lower bound follows from observation that the integrand is monotonically increasing, as shown in \cref{app:strictly_increasing}.
This is a result of our choice of path in \eqref{eq:path}, which allows us to show the derivative of the integrand is equal to the variance of $U'_\beta(\vz)$ under $\pi_\beta(\vz)$ and is therefore non-negative. For equal spacing of the partitions, where $\beta_k = k / K$, we arrive at the \gls{TVO} in \eqref{eq:tvo_lower}, illustrated in \cref{fig:introduction/tvo} (middle). We present a generalized variant with non-equal spacing in \cref{app:generalized_tvo}.

Maximizing the $\ELBO(\theta, \phi, \vx)$ can be seen as a special case of the \gls{TVO}, since for $\beta = 0$, $\pi_\beta(\vz) = q_\phi(\vz \given \vx)$, and so the integrand in \eqref{eq:tvo/tv_identity} becomes $\E_{q_\phi(\vz \given \vx)}\left[\log \frac{p_{\theta}(\vx,\vz)}{q_{\phi}(\vz | \vx)}\right]$, which is equivalent to the definition of \gls{ELBO} in \eqref{eq:introduction/elbo}.
Because the integrand is increasing, we have
\begin{align}
  \ELBO(\theta, \phi, \vx) \leq \TVO(\theta, \phi, \vx) \leq \log p_\theta(\vx),
\end{align}
which means that the \gls{TVO} is an alternative to \gls{IWAE} for tightening the variational bounds. In \cref{app:divergence} we show maximizing the \gls{TVO} is equivalent to minimizing a divergence between the variational distribution and the true posterior $p_{\theta}(\vz | \vx)$.

The integrand in \eqref{eq:tvo/tv_identity} is typically estimated by long running Markov chain Monte Carlo chains computed at different values of $\pi_\beta(\vz)$ \citep{friel2008marginal, lartillot2006computing}.
Instead, we propose a simple importance sampling mechanism that allows us to reuse samples across an arbitrary number of discretizations and which is compatible with gradient-based learning.

%!tex root=./tvo.tex
\section{Optimizing the TVO}
\label{sec:optimization}

We now provide a novel score-function based gradient estimator for the \gls{TVO} which does not require the reparameterization trick.

\paragraph{Gradients}
To use the \gls{TVO} as a variational objective we must be able to differentiate through terms of the form $\nabla_\lambda \E_{\pi_{\lambda, \beta}}[f_\lambda(\vz)]$, where both $\pi_{\lambda, \beta}(\vz)$ and $f_\lambda(\vz) $ are parameterized by $\lambda$, and $\pi_{\lambda, \beta}(\vz)$ contains an intractable normalizing constant.
In the \gls{TVO}, $f_\lambda(\vz)$ is the instantaneous \textsc{elbo} and $\lambda := \{\theta, \phi\}$, but our method is applicable for generic $f_\lambda(\vz):\mathbb{R}^M \mapsto \mathbb{R}$.

We can compute such terms using the \textit{covariance gradient estimator} (derived in \cref{app:gradient_estimator_derivation})
\begin{align}
  \nabla_\lambda \E_{\pi_{\lambda, \beta}}[f_\lambda(\vz)]
  = \E_{\pi_{\lambda, \beta}}[\nabla_\lambda f_\lambda(\vz)] + \Cov_{\pi_{\lambda, \beta}}\left[\nabla_\lambda \log \tilde{\pi}_{\lambda,\beta}(\vz), f_\lambda(\vz)\right] \label{eq:method/estimation/grad}
\end{align}
We emphasize that, like \textsc{reinforce}, our estimator relies on the log-derivative trick, but crucially \textit{unlike} \textsc{reinforce}, doesn't require differentiating through the normalizing constant $Z_\beta = \int \tilde{\pi}_{\lambda,\beta}(\vz) \vdz$.  We clarify the relationship between our estimator and \textsc{reinforce} in \cref{app:covariance_estimator_properties}.

The covariance in \eqref{eq:method/estimation/grad} has the same dimensionality as $\lambda \in \R^D$ because it is between $\nabla_{\lambda} \log \tilde{\pi}_{\lambda,\beta}(\vz) \in \mathbb{R}^{D}$ and $f_\lambda(\vz) \in \R$ and is defined as
\begin{align}
\Cov_{\pi_{\lambda, \beta}}(\mathbf{a}, b) := \E_{\pi_{\lambda, \beta}}\left[(\mathbf{a} - \E_{\pi_{\lambda, \beta}}[\mathbf{a}])(b - \E_{\pi_{\lambda, \beta}}[b])\right].
\end{align}
To estimate this, we first estimate the inner expectations which are then used in estimating the outer expectation.
Thus, estimating the gradient in \eqref{eq:method/estimation/grad} requires estimating expectations under $\pi_\beta$.

\paragraph{Expectations}
By using $q_\phi(\vz \given \vx)$ as the proposal distribution in $S$-sample importance sampling, we can estimate an expectation of a general function $f(\vz)$ under any $\pi_{\beta}(\vz)$ by simply raising each unnormalized importance weight to the power $\beta$ and normalizing:
\begin{align}
    \E_{\pi_{\beta}}[f(\vz)] \approx \sum_{s = 1}^S \overline{w_s^{\beta}} f(\vz_s), \label{eq:method/estimation/general_functions}
\end{align}
where $\vz_s \sim q_\phi(\vz \given \vx)$, $\overline{w_s^{\beta}} := w_s^\beta / \sum_{s' = 1}^S w_{s'}^\beta$ and $w_s := \frac{p_\theta(\vx, \vz_s)}{q_\phi(\vz_s \given \vx)}$. This follows because each unnormalized importance weight can be expressed as
\begin{align}
    \frac{\tilde \pi_\beta(\vx, \vz_s)}{q_\phi(\vz_s \given \vx)}
    = \frac{p_\theta(\vx, \vz_s)^\beta q_\phi(\vz_s \given \vx)^{1 - \beta}}{q_\phi(\vz_s \given \vx)}
    = \frac{p_\theta(\vx, \vz_s)^\beta}{ q_\phi(\vz_s \given \vx)^\beta}
    = \left(\frac{p_\theta(\vx, \vz_s)}{q_\phi(\vz_s \given \vx)}\right)^{\beta}
    = w_s^{\beta}.
\end{align}
Instead of sampling $SK$ times, we can reuse $S$ samples $\vz_s \sim q_\phi(\vz \given \vx)$ across an arbitrary number of terms, since evaluating the normalized weight $\overline{w_s^{\beta_k}}$ only requires raising each weight to different powers of $\beta_k$ before normalizing.
Reusing samples in this way is a use of the method known as ``common random numbers'' and we include experimental results showing it reduces the variance of the covariance estimator in \cref{app:covariance_estimator_properties}~\citep{owen2013monte}.

The covariance estimator does not require $\vz$ to be reparameterizable, which means it can be used in the cases of both non-reparameterizable continuous latent variables and discrete latent variables (without modifying the model using continuous relaxations~\citep{jang2017categorical,maddison2017concrete}).

%!tex root=./tvo.tex
\section{Generalizing Variational Objectives}
\label{sec:method/generalizations}
As previously observed, the left single Riemann approximation of the \gls{TVI} equals the \gls{ELBO}, while the right endpoint ($\beta=1$) is equal to the \gls{EUBO}. The \gls{EUBO} is  analogous to the \gls{ELBO} but under the true posterior and is defined
\begin{align}
    \EUBO(\theta, \phi, \vx) := \E_{p_{\theta}(\vz|\vx)}\left[\log \frac{p_{\theta}(\vx, \vz)}{q_{\phi}(\vz | \vx)}\right].
\end{align}
We also have the following identity
\begin{align}
    \EUBO(\vx, \theta, \phi) &= \log p_\theta(\vx) + \KL{p_\theta(\vz \given \vx)}{q_\phi(\vz \given \vx)}
\end{align}
which should be contrasted against \eqref{eq:introduction/elbo}. We define an upper-bound variant of the \gls{TVO} using the right (rather than left) Riemann sum. Setting $\beta_k = k / K$
\begin{align}
    \TVO_K^U(\theta, \phi, \vx) := \frac{1}{K} \left[\EUBO(\theta, \phi, \vx) + \sum_{k=1}^{K-1}\E_{\pi_{\beta_{k}}}\left[\log \frac{p_{\theta}(\vx, \vz)}{q_{\phi}(\vz | \vx)}\right]\right] \ge \log p(\vx). \label{eq:tvo_upper_gen}
\end{align}

The \gls{WS}~\citep{hinton1995wake} and \gls{RWS}~\citep{bornschein2015reweighted} algorithms have traditionally been viewed as using different objectives during the wake and sleep phase.
The endpoints of the \gls{TVI}, which the \gls{TVO} approximates, correspond to the two objectives used in wake-sleep. We can therefore view \gls{WS} as alternating between between $\TVO_1^L$ and $\TVO_1^U$, i.e. a left and right single term Riemann approximation to the \gls{TVI}. We show this algebraically in \cref{app:special_cases} and additionally, show how the objectives used in variational inference~\citep{blei2017variational}, variational autoencoders~\citep{kingma2014auto,rezende2014stochastic}, and inference compilation~\citep{le2016inference} are all special cases of $\TVO_1^L$ and $\TVO_1^U$. We refer the reader to~\citep{le2018revisiting} for a further discussion of the wake-sleep algorithm.

%!tex root=./tvo.tex
\section{Related Work}

Thermodynamic integration was originally developed in physics to calculate the difference in free energy of two molecular systems~\citep{evans1986molecular}.
\citet{neal1993probabilistic} and \citet{gelman1998simulating} then introduced  \gls{TI} into the statistics community  to calculate the ratios of normalizing constants of general probability models.
\gls{TI} is now commonly used in phylogenetics to calculate the Bayes factor $B = p(x|M_1)/p(x|M_0)$, where $M_0, M_1$ are two models specifying (for instance) tree topologies and branch lengths~\citep{lartillot2006computing, xie2010improving, rodrigue2011fast}.
We took inspiration from~\citet{fan2010choosing} who replaced the ``power posterior'' $p(\theta|\vx, M, \beta)=p(x|\theta, M)^{\beta}p(\theta, M) / Z_{\beta}$ of~\citet{xie2010improving} with $p(\theta|\vx, M, \beta) = [p(\vx|\theta, M)p(\theta|M)]^{\beta}[p_0(\theta|M)]^{1 - \beta} / Z_{\beta}$, where $p_0(\theta|M)$ is a tractable reference distribution chosen to facilitate sampling. That the integrand in~\cref{eq:tvo/tv_identity} is strictly increasing was observed by~\citet{lartillot2006computing}.

We refer the reader to \citet{titsias2018unbiased} for a summary of the numerous advances in variational methods over recent years. The method most similar to our own was proposed by \citet{bornschein2016bidirectional}, who introduced another way of improving deep generative modeling through geometrically interpolating between distributions and using importance sampling to estimate gradients.
Unlike the \gls{TVO}, they define a lower bound on the marginal likelihood of a modified model defined as $(p_\theta(\mathbf x, \mathbf z)q_\phi(\mathbf z \given \mathbf x)q(\mathbf x))^{1/2} / Z$ where $q(\mathbf x)$ is an auxiliary distribution.

\citet{grosse2013annealing} studied \gls{AIS}, a related technique that estimates partition functions using a sequence of intermediate distributions to form a product of ratios of importance weights.
They observe the geometric path taken in \gls{AIS} is equivalent to minimizing a weighted sum of \gls{KL} divergences, and use this insight to motivate an alternative path. To the best of our knowledge, our work is the first to explicitly connect \gls{TI} and \gls{VI}.

%!tex root=./tvo.tex

\section{Experiments}
\label{sec:experiments}

\subsection{Discrete Deep Generative Models}
\label{sec:discrete_experiment}
We use the \gls{TVO} to learn the parameters of a deep generative model with discrete latent variables.\footnote{Code to reproduce all experiments is available at: \url{https://github.com/vmasrani/tvo}.}
We use the binarized MNIST dataset with the standard train/validation/test split of 50k/10k/10k~\citep{salakhutdinov2008quantitative}. We train a sigmoid belief network, described in detail in \cref{app:vae_details}, using the \gls{TVO} with the Adam optimizer.  In the first set of experiments we investigate the effect of the discretization $\beta_{0:K}$, number of partitions $K$ and number of particles $S$. We then compare against \gls{VIMCO} and \gls{RWS} (with the wake-$\phi$ objective) state-of-the-art \gls{IWAE}-based methods for learning discrete latent variable models~\citep{le2018revisiting}. All figures have been smoothed for clarity.

\begin{figure}[t]
    \centering
    \includegraphics[width=\textwidth]{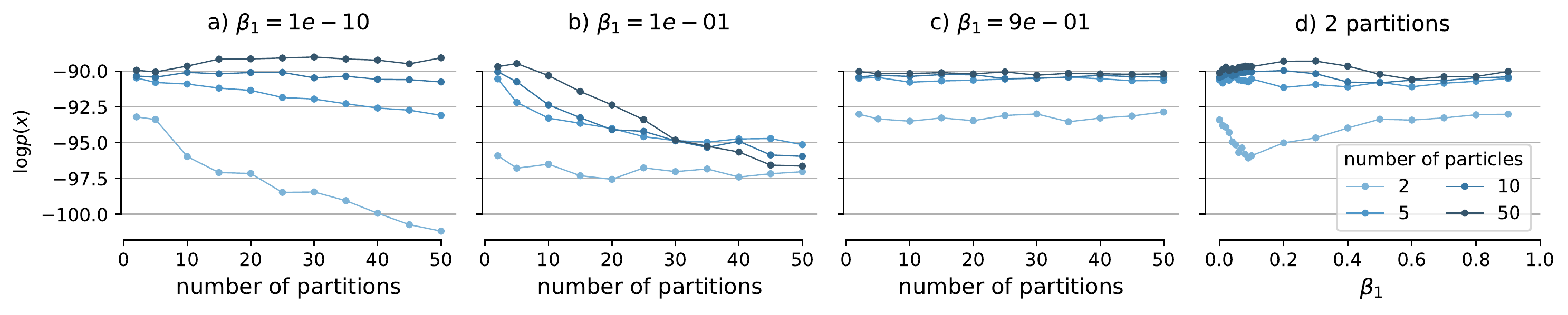}
    \caption{Investigation of how number of particles $S$, number of partitions $K$, and $\beta_1$ affect learning of the generative model.
    In the first three plots (a-c), we vary $S$ and $K$ for different values of $\beta_1$ and observe that while $S$ should be as high as possible, there is an optimal value for $K$, beyond which performance begins to degrade.
    Assuming $\beta_1$ is well-chosen, we see that as few as $K = 2$ partitions can result in good model learning, as seen in the last plot (d).}
    \label{fig:experiments/discrete_vae/insights}
\end{figure}

\begin{figure}[t]
    \centering
    % \savebox{\tempfig}{\includegraphics[width=0.40\textwidth]{figures/discrete_vae/baselines}}% Store larger image in box
    % \raisebox{\dimexpr\ht\tempfig-\height}{\includegraphics[width=0.40\textwidth]{figures/discrete_vae/efficiency}}
    \includegraphics[width=0.49\textwidth]{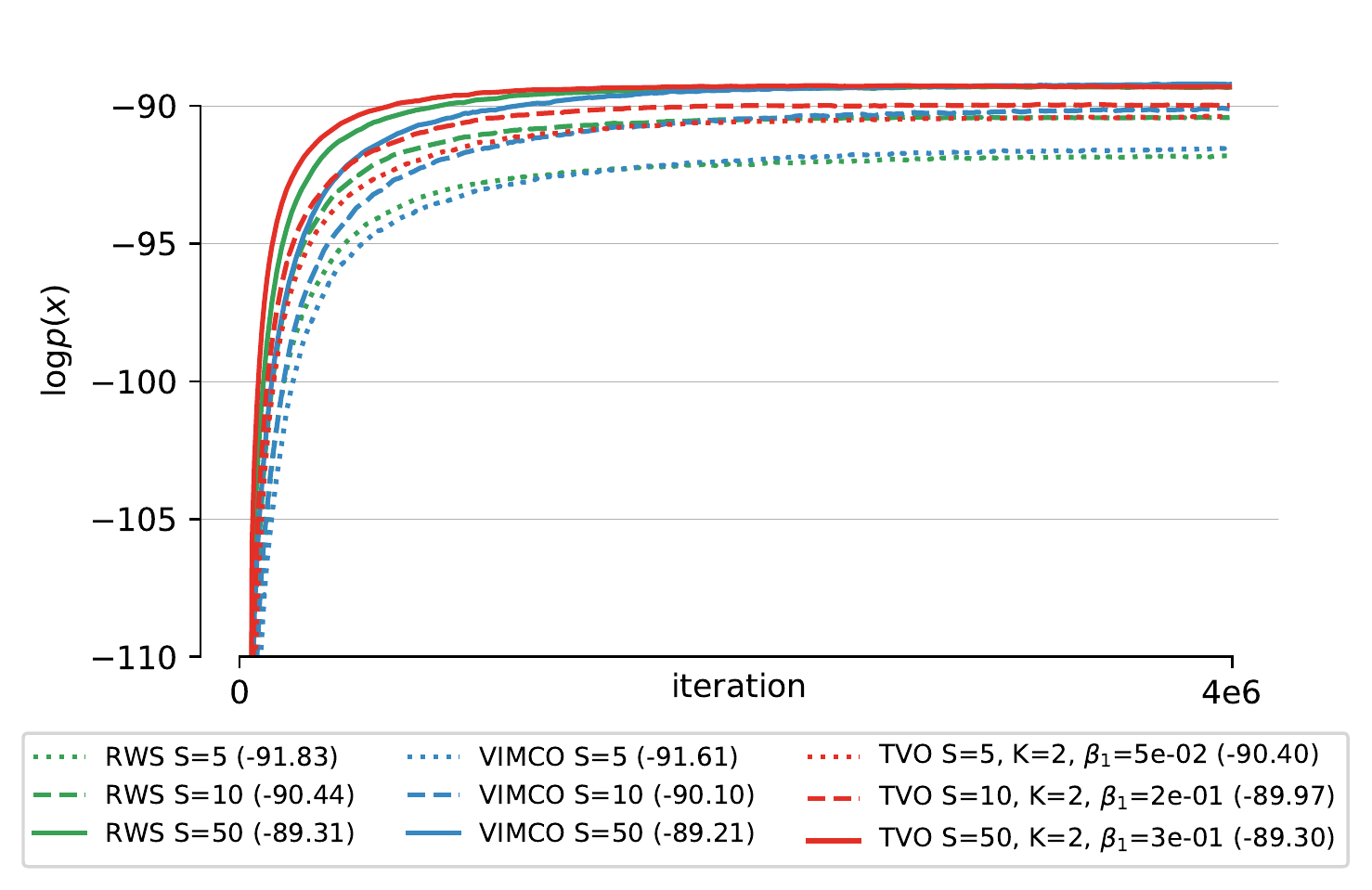}
    \includegraphics[width=0.49\textwidth]{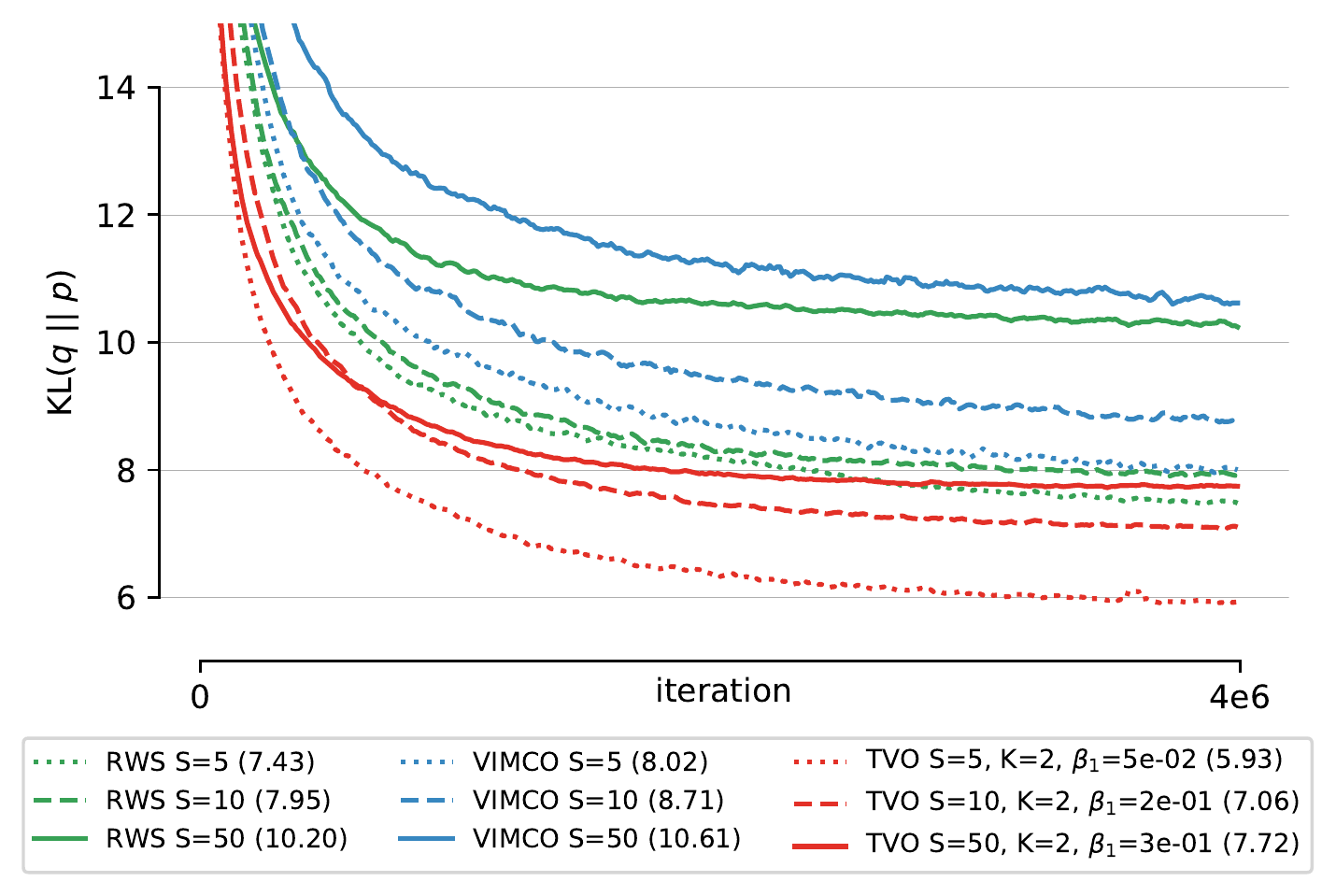}
    \caption{Comparisons with baselines on a held out test set. \emph{(Left)} Learning curves for different methods. For \gls{TVO} outperforms other methods both in terms of speed of convergence and the learned model for $S<50$. At $S=50$ \textsc{vimco} achieves a higher test log evidence but takes longer to converge than the $\gls{TVO}$. \emph{(Right)} \gls{KL} divergence between current $q$ and $p$ (which measures how well $q$ ``tracks'' $p$) is lowest for \gls{TVO}.}
    \label{fig:experiments/discrete_vae/baselines}
\end{figure}

\textbf{The effect of $S$, $K$, and $\beta$ locations}
We expect that increasing the number of partitions $K$ makes the Riemann sum approximate the integral over $\beta$ more tightly.
However, with each addition term, we add noise due to the use of importance sampling to estimate the expectation $\E_{\pi_{\beta}}[\log p/q]$.
Importance sampling estimates of points on the curve further to the right are likely to be more biased because $\pi_{\beta}$ gets further from $q$ as we increase $\beta$.
We found the combination of these two effects means that there is a ``sweet spot,'' or an optimal number of partitions beyond which adding more partitions becomes detrimental to performance.

\begin{wrapfigure}{r}{0.27\textwidth}
    \begin{center}
      \includegraphics[width=0.27\textwidth]{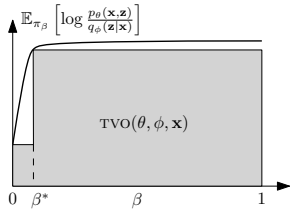}
    \end{center}
    \caption{The location of $\beta^*$, the point of maximum curvature.}
    \label{fig:thermo_diagram/curvature}
\end{wrapfigure}

We have empirically observed that the curve in~\cref{fig:introduction/tvo} is often rising sharply from $\beta = 0$ until a point of maximum curvature $\beta^*$, after which it is almost flat until $\beta = 1$, as seen in~\cref{fig:thermo_diagram/curvature}.
We hypothesized that if $\beta_1$ is located far before $\beta^*$ (the point of maximum curvature), a large number of additional partitions would be needed to capture additional area, while if $\beta_1$ is located after $\beta^*$, additional partitions would simply incur a high cost of bias without significantly tightening the bound.
To investigate this, we choose small ($10^{-10}$), medium ($0.1$) and large ($0.9$) values of $\beta_1$, and logarithmically space the remaining $\beta_{2:K}$ between $\beta_1$ and $1$. For each value of $\beta_1$ we train the discrete generative model for $K \in \{2, 5, 10, \dotsc, 50\}$ and $S \in \{2, 5, 10, 50\}$, and show the test log evidence at the last iteration of each trial, approximated by evaluating the \gls{IWAE} loss with $5000$ samples.

Our hypothesis is corroborated in~\cref{fig:experiments/discrete_vae/insights}, where we observe in \cref{fig:experiments/discrete_vae/insights}a that for $\beta_1 = 10^{-10}$ a large number of partitions are needed to approximate the integral. In \cref{fig:experiments/discrete_vae/insights}b we increase $\beta_1$ to $10^{-1}$ and observe only a few partitions are needed to improve performance, after which adding additional partitions becomes detrimental to model learning.

From \cref{fig:experiments/discrete_vae/insights}c we can see that if $\beta_1$ is chosen to be well beyond $\beta^*$, the Riemann sum cannot recover the ``lost'' area even if the number of partitions is increased.
That the performance does not degrade in this case is due to the fact that for sufficiently high $\beta_k$, the curve in \cref{fig:introduction/tvo} is flat and therefore $\pi_{\beta_k} \approx \pi_{\beta_k+1} \approx p_\theta(\vz|\vx)$. We also observe that increasing number of samples $S$---which decreases importance sampling bias per partition---improves performance in all cases.

In our second experiment, shown in the \cref{fig:experiments/discrete_vae/insights}d, we fix $K = 2$ and investigate the quality of the learned generative model for different $\beta_1$,
This plot clearly shows $\beta^*$ is somewhere near $0.3$, as model learning improves as $\beta_1$ approaches this point then begins to degrade.

Given these results, we recommend using as many particles $S$ as possible and performing a hyper-parameter search over $\beta_1$ (with $K = 2$) when using the \gls{TVO} objective. We leave finding the optimal placement of discretization points to future work.

\textbf{Performance}
In \cref{fig:experiments/discrete_vae/baselines} (left), we compare the \gls{TVO} against \textsc{vimco} and \gls{RWS} with the wake-$\phi$ objective, the state-of-the-art \gls{IWAE}-based methods for learning discrete latent variable models~\citep{le2018revisiting}.
For $S < 50$, the \gls{TVO} outperforms both methods in terms of speed of convergence and the final test log evidence $\log p_\theta(\vx)$, estimated using $5000$ \gls{IWAE} particles as before. At $S = 50$ \textsc{vimco} achieves a higher test log evidence but converges more slowly.

We also investigate the quality of the learned inference network by plotting the \gls{KL} divergence (averaged over the test set) between the current $q$ and current $p$ as training progresses (\cref{fig:experiments/discrete_vae/baselines} (right)).
This indicates how well $q$ ``tracks'' $p$.
This is estimated as log evidence minus \gls{ELBO} where the former is estimated as before and the latter is estimated using $5000$ Monte Carlo samples.
The \gls{KL} is lowest for \gls{TVO}.

Somewhat surprisingly, for all methods, increasing number of particles makes the \gls{KL} worse.
We speculate that this is due to the ``tighter bounds'' effect of \citet{rainforth2018tighter}, who showed that increasing the number of samples can positively affect model learning but adversely affect inference network learning, thereby increasing the \gls{KL} between the two.

\begin{figure}[t]
    \centering
    \includegraphics[width=0.49\textwidth,valign=t]{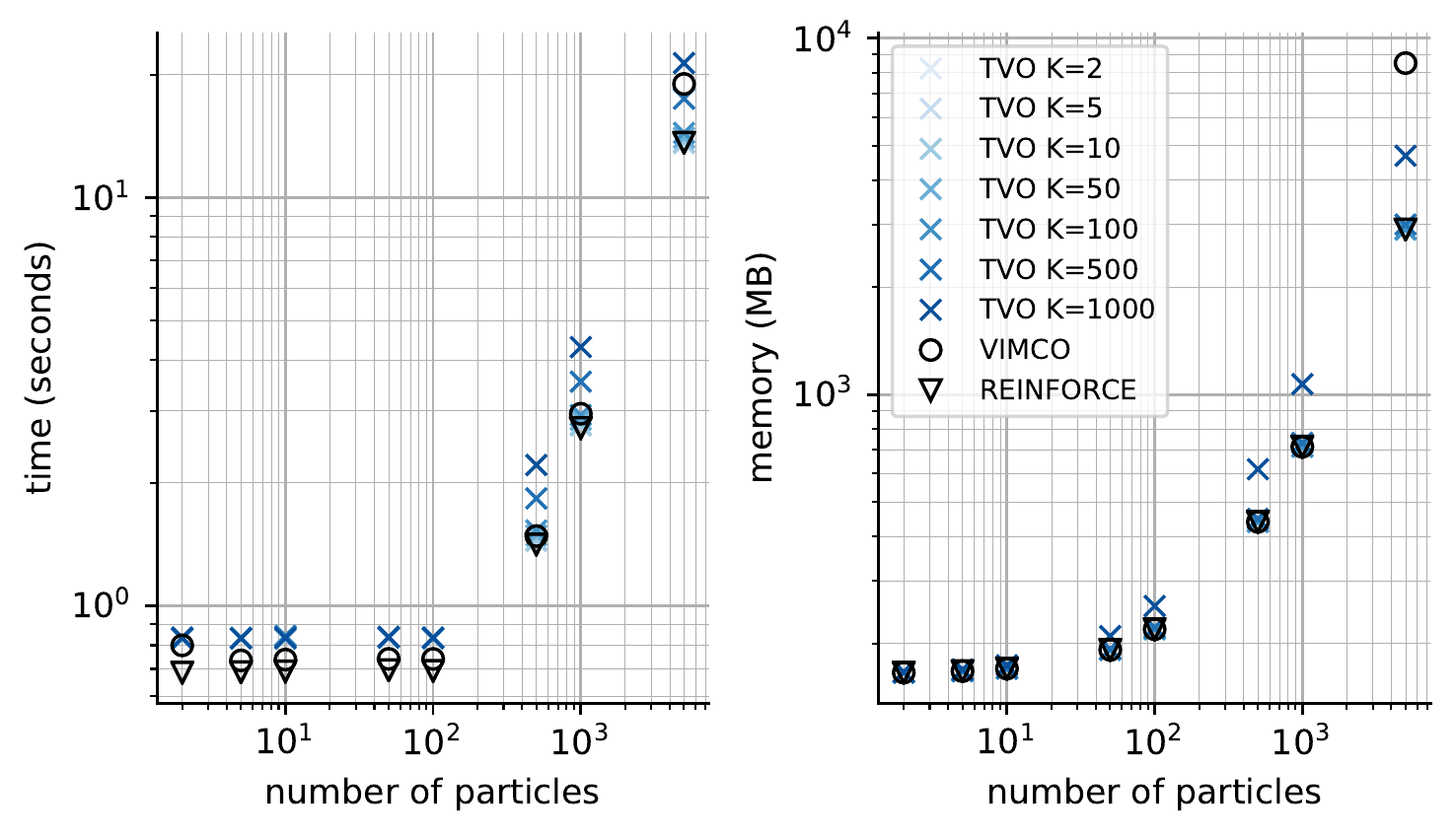}
    \includegraphics[width=0.49\textwidth,valign=t]{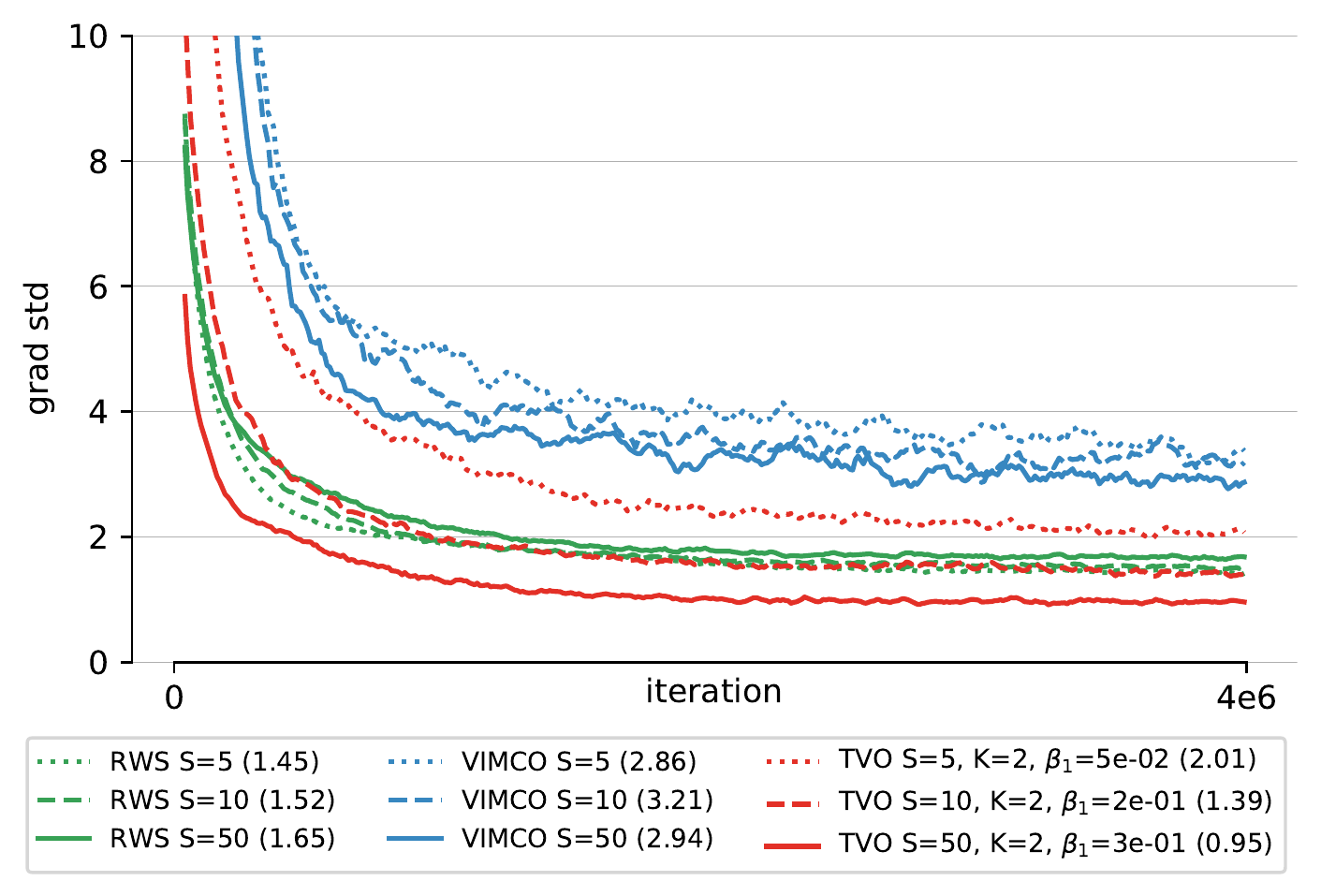}
    \caption{Computational and gradient estimator efficiency. \emph{(Left)} Time and memory efficiency of the \gls{TVO} with increasing number of partitions vs baselines, measured for $100$ iterations of optimization.
    Increasing the number of partitions is much cheaper than increasing the number of particles.
    \emph{(Right)} Standard deviation of the gradient estimator for each objective. \gls{TVO} is lowest variance, \textsc{vimco} is highest variance, \gls{RWS} is in the middle.}
    \label{fig:experiments/discrete_vae/kl_and_grad_std}
\end{figure}

\textbf{Efficiency}
Since we use $K = 2$ partitions for the same number of particles $S$, the time and memory complexity of \gls{TVO} is double that of other methods.
While this is true, in both time and memory cases, the constant factor for increasing $S$ is much higher than for increasing $K$.
As shown in \cref{fig:experiments/discrete_vae/kl_and_grad_std} (left), it is virtually free to increase number of partitions.
This is because for each new particle, we must additionally sample from the inference network and score the sample under both $p$ and $q$ to obtain a weight.
On the other hand, we can reuse the $S$ samples and corresponding weights in estimating values for the $K + 1$ terms in the Riemann sum.
Thus, the region of the the computation graph that is dependent on $K$ is \textit{after} the expensive sampling and scoring, and only involves performing basic operations on additional matrices of size $S \times K$.

\textbf{Variance}
In \cref{fig:experiments/discrete_vae/kl_and_grad_std} (right), we plot the standard deviation of the gradient estimator for each method, where we compute the standard deviation for the $d$\textsuperscript{th} element of the gradient estimated over 10 samples and take the average across all $D$.

The gradient estimator of the \gls{TVO} has lower variance than both \textsc{vimco}, which uses \textsc{reinforce} with a control variate as a gradient estimator and \gls{RWS} which can calculate the gradient without reparameterizing or using the log-derivative trick.
At $S = 5$, \gls{RWS} has lower gradient variance but its performance is worse in terms of both model and inference learning.

% The gradient estimator of the \gls{TVO} has lower variance than \textsc{vimco}, which uses \textsc{reinforce} with a control variate as a gradient estimator. As expected, the variance of the \gls{TVO} is higher than \textsc{rws}, which can calculate the gradient without reparameterizing or using the log-derivative trick.
% This indicates that \gls{TVO} is a new, potentially advantageous point on the bias-variance trade-off curve along with \textsc{reinforce} (high variance, zero bias), \textsc{vimco} (lower variance, zero bias), \textsc{rws} (even lower variance, biased).

\subsection{Continuous Deep Generative Models}
\label{sec:continuous_experiment}
Using the binarized MNIST dataset and experimental design described above, we also evaluated our method on a deep generative model with continuous latent variables. The model is described in detail in \cref{app:vae_details}.
For each $S \in \{5, 10, 50\}$ we sweep over $K \in \{2, ..., 6\}$ and $20$ $\beta_1$ values linearly spaced between $10^{-2}$ and $0.9$. We optimize the objectives using the Adam optimizer with default parameters.

\begin{figure}[t]
    \begin{minipage}{0.49\textwidth}
      \centering
      \includegraphics[width=\textwidth]{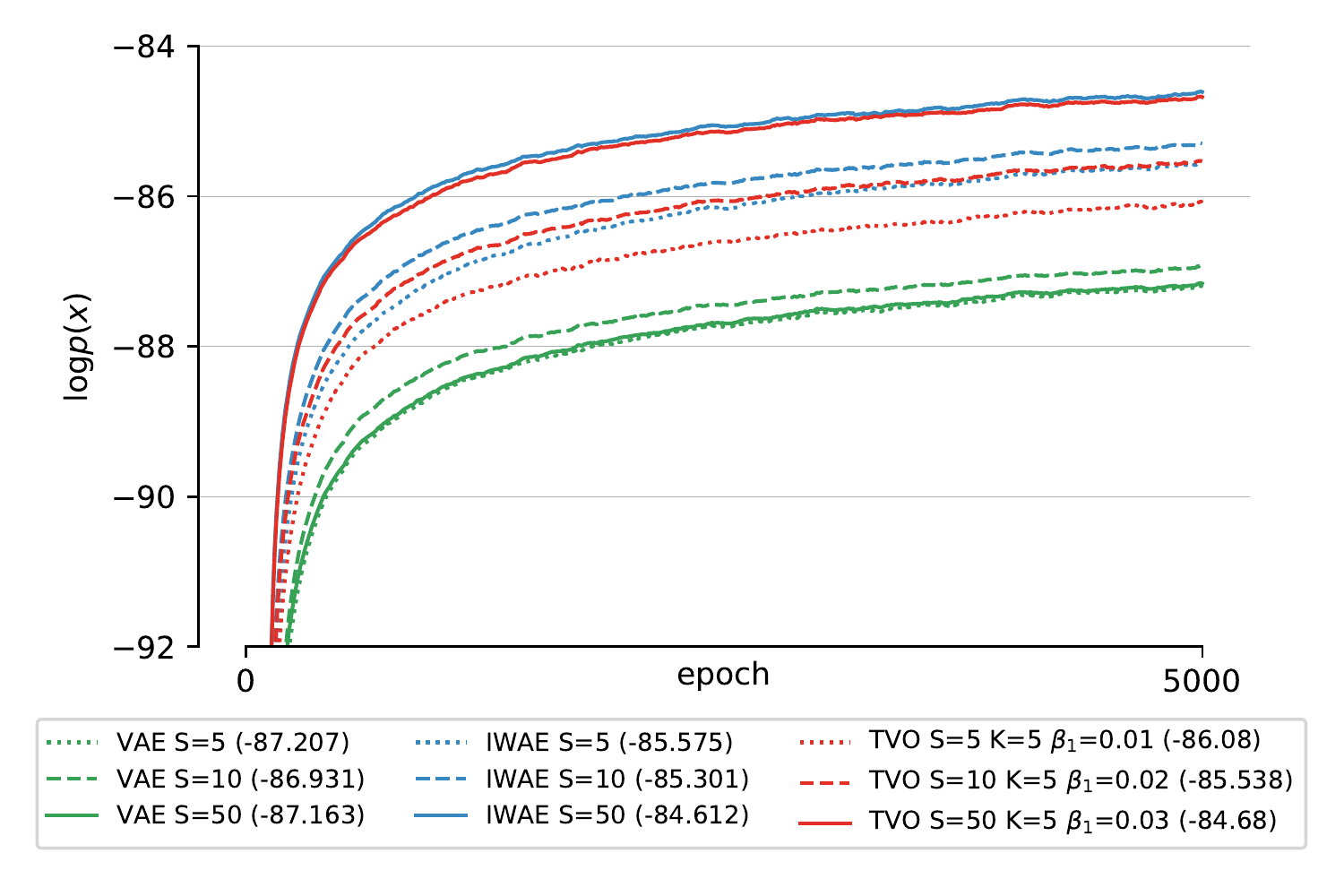}
    \end{minipage}\hfill
    \begin{minipage}{0.49\textwidth}
      \centering
      \includegraphics[width=\textwidth]{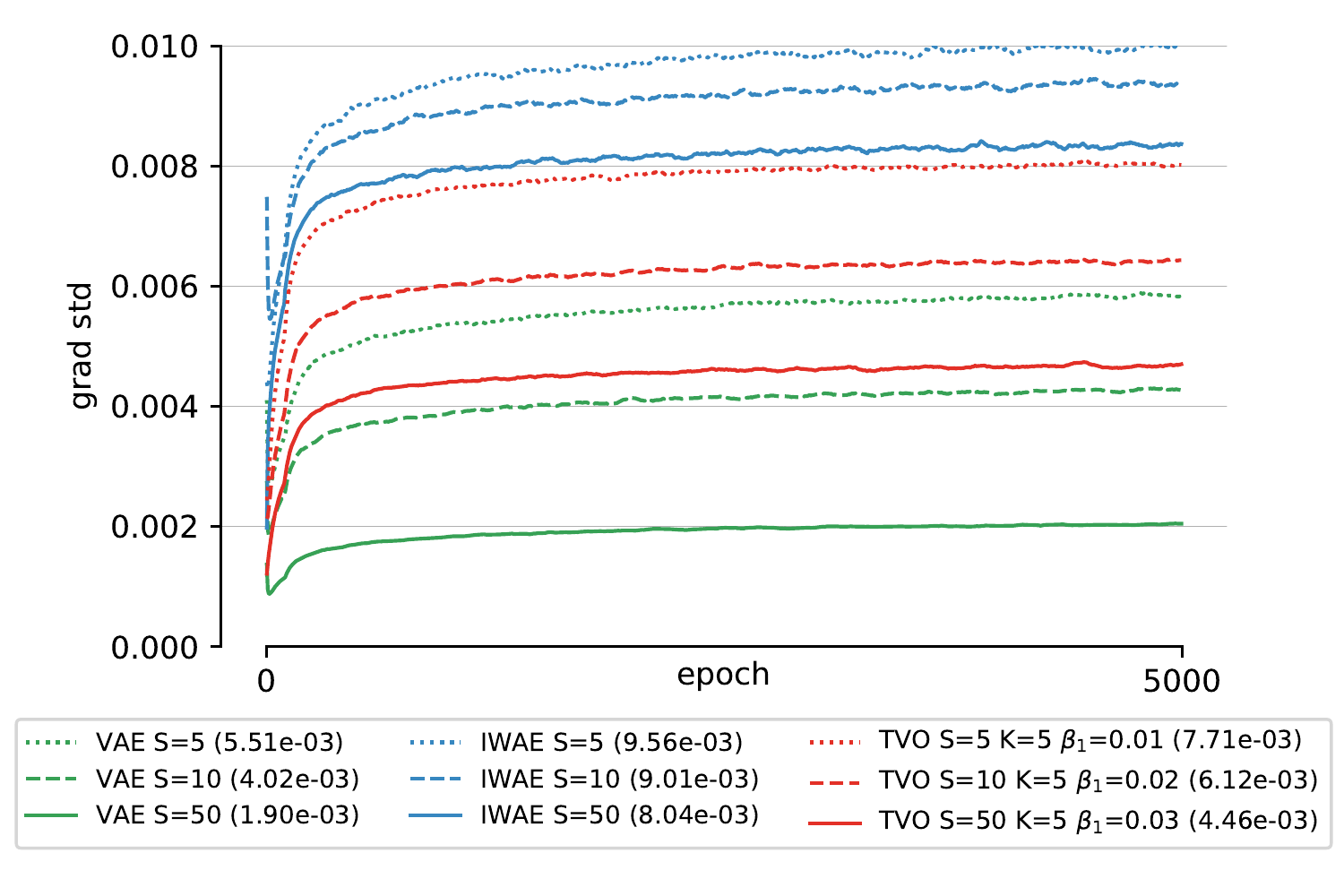}
    \end{minipage}
    \caption{Learning curves for learning continuous deep generative models using different objectives.
    \emph{(Left)}
        Despite not using the reparameterization trick, \gls{TVO} outperforms \glspl{VAE} and is competitive with \gls{IWAE} at 50 samples. For all S, \gls{IWAE} $>$ \gls{TVO} $>$ \gls{VAE}.
    \emph{(Right)}
        Standard deviation of the gradient estimator for each objective. The \gls{TVO} has lower variance than \gls{IWAE} but higher than \gls{VAE}.
    }
    \label{fig:experiments/continuous_vae/baselines}
 \end{figure}

\textbf{Performance}
In \cref{fig:experiments/continuous_vae/baselines} (left), we train the model using the \gls{TVO} and compare against the same model trained using the single sample \gls{VAE} objective and multisample \gls{IWAE} objective. The \gls{TVO} outperforms the \gls{VAE} and performs competitively with \gls{IWAE} at 50 samples, despite not using the reparameterization trick. \gls{IWAE} is the top performing objective in all cases. As in the discrete case, increasing the number of particles $S$ improves model learning for all methods, but the improvement is most significant for the \gls{TVO}. Interestingly \gls{VAE} performance actually \textit{decreases} when the number of samples increases from 10 to 50. A similar effect was noticed by \citet{burda2016importance} on the omniglot dataset.

\textbf{Variance}
In \cref{fig:experiments/continuous_vae/baselines} (right), we plot the standard deviation of each method's gradient estimator. The standard deviation of the \gls{TVO} estimator falls squarely between \gls{VAE} (best) and \gls{IWAE} (worst). The variance of each method improves as the number of samples increases, and as in the discrete model, the improvement is most significant in the case of \gls{TVO}. Unlike in the discrete case, the variance does not decrease as the optimization proceeds, but plateaus early and then gradually increases. In \cref{app:covariance_estimator_properties} we include additional experiments to evaluate the properties of the covariance gradient estimator when used on the \gls{ELBO}.

For both \gls{IWAE} and the \gls{TVO}, increasing the number of samples leads to decreased gradient variance and improved model learning.
However, \gls{IWAE} has the best performance but the highest variance across the three models.
These results lend support to the conclusions of~\citet{rainforth2018tighter} who observe that the variance of a gradient estimator ``is not always a good barometer for the effectiveness of a gradient estimation scheme.''

% !TEX root =  tvo.tex
\section{Conclusions}
The thermodynamic variational objective represents a new way to tighten evidence bounds and is based on a tight connection between variational inference and thermodynamic integration.
We demonstrated that optimizing the \gls{TVO} can have a positive impact on the learning of discrete deep generative models and can perform as well as using the reparameterization trick to learn continuous deep generative models.
% This is a subtle point worth emphasizing.
% Tightening evidences bound seems to nearly universally help model learning, but gradients estimated using evidence bounds from the importance sampling family seem to negatively impact the inference network.
% In the end the importance sampling family gradients are not only higher variance but also must be, according to our experimental results, subtly biased so as to lead to less optimal final models.
% Put another way, the joint optimization path provided by the \gls{TVO}-derived gradients seems to implicitly regularize towards better models.
% This and the near equivalence of learning using \gls{TVO} gradients without reparameterization to \gls{IWAE} with reparameterization in deep continuous latent variable models encourages much closer examination and experimentation.

The weakness of our method lies in choosing the discretization points.
This does, however, point out opportunities for future work wherein we adaptively select optimal positions of the $\beta_{1:K}$ points, perhaps using techniques from the Bayesian numerical quadrature literature \citep{o1991bayes,rasmussen2003bayesian,osborne2012active}.

The approximate path integration perspective provided by our development of the \gls{TVO} also sheds light on the connection between otherwise disparate deep generative model learning techniques.
In particular, the \gls{TVO} integration perspective points to ways to improve wake-sleep via tightening the \gls{EUBO} using similar integral upper-bounding techniques.
Further experimentation is warranted to explore how \gls{TVO} insights can be applied to all special cases of the \gls{TVO} including non-amortized variational inference and to the use of the \gls{TVO} as a compliment to annealing importance sampling for final model evidence evaluation.
% Our unreported initial experiments along these lines show small but positive effects which we are currently exploring.

\subsubsection*{Acknowledgments}
We would like to thank Trevor Campbell, Adam {\'S}cibior, Boyan Beronov, and Saifuddin Syed for their helpful comments on early drafts of this manuscript.
Tuan Anh Le's research leading to these results is supported by EPSRC DTA and Google (project code DF6700) studentships.
We acknowledge the support of the Natural Sciences and Engineering Research Council of Canada (NSERC), the Canada CIFAR AI Chairs Program, Compute Canada, Intel, and DARPA under its D3M and LWLL programs.

\small

\bibliography{tvo}

\newpage

\normalsize
\appendix
%!tex root=./tvo.tex

\section{Thermodynamic Integration}
\label{app:TI}
\Glsfirst{TI} is a technique used in physics and phylogenetics to approximate intractable normalized constants of high dimensional distributions~\citep{neal1993probabilistic, gelman1998simulating}.
It is based on the observation that it is easier to calculate the ratio of two unknown normalizing constants than it is to calculate the constants themselves. More formally, consider two densities over space $\mathcal{Z}$
\begin{align}
  \pi_i(\vz) &= \frac{\tilde{\pi}_i(\vz)}{Z_i},
   && Z_i = \int_{\mathcal{Z}} \tilde{\pi}(\vz) \vdz,
    && i \in \{0, 1\}.
\end{align}
To apply TI, we form a continuous family (or ``path'') between $\pi_0(\vz)$ and $\pi_1(\vz)$ via a scalar parameter $\beta \in [0,1]$
\begin{align}
  \pi_{\beta}(\vz) &= \frac{\tilde{\pi}_{\beta}(\vz)}{Z_{\beta}} = \frac{\tilde{\pi}_1(\vz)^{\beta}\tilde{\pi}_0(\vz)^{{1 - \beta}}}{Z_{\beta}},
   && Z_{\beta} = \int_{\mathcal{Z}} \tilde{\pi}_{\beta}(\vz) \vdz,
    && \beta \in [0, 1].  \label{eq:pi_z}
\end{align}
The central identity that allows us to compute the ratio $\log(Z_1/Z_0)$ is derived as follows. Assuming we can exchange integration with differentiation,
\begin{align}
  \frac{\partial \log Z_{\beta}}{\partial \beta} &= \frac{1}{Z_{\beta}} \frac{\partial}{\partial \beta}Z_{\beta} \nonumber \\
  &= \frac{1}{Z_{\beta}} \frac{\partial}{\partial \beta} \int \tilde{\pi}_{\beta}(\vz) \vdz \nonumber \\
  &= \int \frac{1}{Z_{\beta}} \frac{\partial}{\partial \beta} \tilde{\pi}_{\beta}(\vz) \vdz \nonumber \\
  &= \int \frac{\tilde{\pi}_{\beta}(\vz)}{Z_{\beta}}\frac{\partial}{\partial \beta} \log \tilde{\pi}_{\beta}(\vz) \vdz \nonumber,
\end{align}
which directly implies
\begin{align}
  \frac{\partial \log Z_{\beta}}{\partial \beta} &= \E_{\pi_{\beta}}\big[U^{\prime}_{\beta}(\vz)\big], \label{eq:background/thermo/dZdbeta}
\end{align}
where the quantity $U_{\beta}(\vz) = \log \tilde{\pi}_\beta(\vz)$ is referred to as the ``potential'' in statistical physics and $U^{\prime}_{\beta}(\vz) := \frac{\partial}{\partial \beta} U_{\beta}(\vz)$.
The variable $\beta$ can be interpreted as the inverse temperature parameter.
Because one can typically compute $\log \tilde{\pi}_\beta(\vz)$, \eqref{eq:background/thermo/dZdbeta} allows us to exchange the first derivative of something we cannot compute with an expectation over something we can compute.
Then, to calculate the ratio $\log(Z_1/Z_0)$ we integrate out $\beta$ on both sides of \eqref{eq:background/thermo/dZdbeta}
\begin{align}
\int_0^1 \frac{\partial \log Z_\beta}{\partial \beta} d\beta &= \int_0^1 \E_{\pi_{\beta}}\big[U^{\prime}_{\beta}(\vz)\big] d\beta
\end{align}
which via the fundamental theorem of calculus results in
\begin{align}
  \log(Z_1) - \log(Z_0) &= \int_0^1 \E_{\pi_{\beta}}\big[U^{\prime}_{\beta}(\vz)\big] d\beta.
\end{align}

\section{The Increasing Integrand}
\label{app:strictly_increasing}

\subsection{Notation}
\label{app:integrand_notation}
\begin{align}
    \log p_\theta(\vx) &= \int_0^1 g(\beta) d\beta\\
    g(\beta) &= \E_{\pi_\beta(\vz)}\left[ U^{\prime}(\vz) \right]\\
    U^{\prime}(\vz) &= \log \frac{p_\theta(\vx, \vz)}{q_\phi(\vz | \vx)}
\end{align}
Given our choice of geometric path $\pi_\beta(\vz) = \tilde{\pi}_{\beta}(\vz) / Z_{\beta}$, $\tilde{\pi}_{\beta}(\vz) = p(\vx, \vz)^{\beta}q(\vz|\vx)^{1-\beta}$, the potential $U^{\prime}(\vz) = \frac{\partial}{\partial \beta} \log \tilde{\pi}_{\beta}(\vz)$ loses its dependency on $\beta$ after differentiating. This allows us to show
\begin{align}
    \frac{\partial}{\partial \beta} g(\beta) = \Var_{\pi_{\beta}(\vz)}[U^{\prime}(\vz)]\label{eq:gvar_equivalence}
\end{align}
which means $\frac{\partial}{\partial \beta} g(\beta) \ge 0, \forall \beta \in [0, 1]$ and therefore that $g(\beta)$ is monotonically non-decreasing. Changes between lines are tracked in \textcolor{blue}{blue}.

\begin{proof}[Proof of Equation~\eqref{eq:gvar_equivalence}]
\begin{align*}
    \frac{\partial}{\partial \beta} g(\beta) &= \frac{\partial}{\partial \beta} \E_{\pi_\beta(\vz)}\left[ U^{\prime}(\vz) \right] \\
    &= \frac{\partial}{\partial \beta} \textcolor{blue}{\bigg[ \int \pi_{\beta}(\vz) U^{\prime}(\vz) \vdz]\bigg]}\\
    &= \int U^{\prime}(\vz) \textcolor{blue}{\frac{\partial}{\partial \beta} \pi_{\beta}(\vz)} \vdz \\
    &= \int U^{\prime}(\vz) \frac{\partial}{\partial \beta}\big[\textcolor{blue}{Z_{\beta}^{-1} \tilde{\pi}_{\beta}(\vz)} \big] \vdz\\
    &= \int U^{\prime}(\vz)\bigg[\textcolor{blue}{\tilde{\pi}_{\beta}(\vz)\frac{\partial}{\partial \beta}Z_{\beta}^{-1} + Z_{\beta}^{-1}\frac{\partial}{\partial \beta}\tilde{\pi}_{\beta}(\vz)}\bigg] \vdz.
\end{align*}

Now we use the ``inverse log-derivative'' trick $\frac{\partial}{\partial x}(f(x)^{-1}) = -\frac{1}{f(x)} \frac{\partial}{\partial x}\log f(x)$ on the first term, and the log-derivative trick on the second
\begin{align}
    &= \int U^{\prime}(\vz)\bigg[ \tilde{\pi}_{\beta}(\vz)\textcolor{blue}{\frac{-1}{Z_{\beta}}\frac{\partial}{\partial \beta}\log Z_{\beta}} + \frac{1}{Z_{\beta}}\textcolor{blue}{\tilde{\pi}_{\beta}(\vz) \frac{\partial}{\partial \beta}\log \tilde{\pi}_{\beta}(\vz)}\bigg] \vdz\\
    &= \int U^{\prime}(\vz)\left[-\textcolor{blue}{\pi_{\beta}(\vz)} \frac{\partial}{\partial \beta}\log Z_{\beta} + \textcolor{blue}{\pi_{\beta}(\vz)} \frac{\partial}{\partial \beta}\log \tilde{\pi}_{\beta}(\vz)\right] \vdz, \label{eq:convex_e3}
\end{align}
Then we use \eqref{eq:background/thermo/dZdbeta} on the first term, and the definition of $U^{\prime}(\vz)$ in the second
\begin{align}
    &= \int U^{\prime}(\vz)\bigg[-\pi_{\beta}(\vz)\textcolor{blue}{\E_{\pi_{\beta}(\vz)}\big[U^{\prime}(\vz)\big]} + \pi_{\beta}(\vz) \textcolor{blue}{U^{\prime}(\vz)} \bigg] \vdz\\
    &= \textcolor{blue}{-\int \pi_{\beta}(\vz)U^{\prime}(\vz)} \E_{\pi_{\beta}(\vz)}\big[U^{\prime}(\vz)\big] \textcolor{blue}{\vdz} + \textcolor{blue}{\int U^{\prime}(\vz)U^{\prime}(\vz)} \pi_{\beta}(\vz) \textcolor{blue}{\vdz}
\end{align}

% \textcolor{blue}{}

Finally we rearrange, noting that the expectation is a scalar and can therefore come out of the integrand
\begin{align}
    &= \textcolor{blue}{-\bigg[\E_{\pi_{\beta}(\vz)}\big[U^{\prime}(\vz)}\big]\bigg] \bigg[\int \pi_{\beta}(\vz)U^{\prime}(\vz) \vdz \bigg] + \int U^{\prime}(\vz) U^{\prime}(\vz) \pi_{\beta}(\vz) \vdz \\
    &= -\big[\E_{\pi_{\beta}(\vz)}[U^{\prime}(\vz)]\big]^{\textcolor{blue}{2}} + \textcolor{blue}{\E_{\pi_{\beta}(\vz)}}\big[U^{\prime}(\vz)^2\big] \\
    &= \textcolor{blue}{\Var_{\pi_{\beta}(\vz)}}[U^{\prime}(\vz)].
\end{align}

Therefore,
\begin{align}
    \frac{\partial}{\partial \beta} g(\beta) &= \Var_{\pi_{\beta}(\vz)}[U^{\prime}(\vz)].
\end{align}
\end{proof}

\section{The generalized TVO}
\label{app:generalized_tvo}
The \gls{TVO} presented in~\cref{sec:method/tvo} is a lower bound to $\log p_\theta(\vx)$ using a left Riemann sum approximation to the thermodynamic variational identity.
Using the right Riemann sum results in an upper bound which can be minimized (rather than maximized) during optimization (cf.~\cref{sec:method/generalizations}).
This loss is used in the inference compilation and during the sleep-phase $\phi$ update in the Wake-Sleep algorithm. Below we present both the upper-bound and lower-bound variants of the \gls{TVO}, with non-equally spaced partitions $0 = \beta_{0} < \beta_{1} < \cdots < \beta_{K} = 1$, $\Delta_{\beta_{k}} = \beta_{k} - \beta_{k-1}$, $K > 1$

\begin{align}
    &\TVO_K^L(\theta, \phi, \vx) := \Delta_{\beta_1}\ELBO(\theta, \phi, \vx) + \sum_{k=2}^{K}\Delta_{\beta_k}\E_{\pi_{\beta_{k-1}}}\left[\log \frac{p_{\theta}(\vx, \vz)}{q_{\phi}(\vz | \vx)}\right] \le \log p(\vx)\label{eq:tvo_lower_gen}\\
    &\TVO_K^U(\theta, \phi, \vx) := \Delta_{\beta_K}\EUBO(\theta, \phi, \vx) + \sum_{k=1}^{K-1}\Delta_{\beta_k}\E_{\pi_{\beta_{k}}}\left[\log \frac{p_{\theta}(\vx, \vz)}{q_{\phi}(\vz | \vx)}\right] \ge \log p(\vx), \label{eq:tvo_upper_gen_2}
\end{align}
where
\begin{align}
    &\ELBO(\theta, \phi, \vx) := \E_{q_{\phi}(\vz | \vx)}\left[\frac{p_{\theta}(\vx, \vz)}{q_{\phi}(\vz | \vx)}\right],&&
    \EUBO(\theta, \phi, \vx) := \E_{p_{\theta}(\vz|\vx)}\left[\frac{p_{\theta}(\vx, \vz)}{q_{\phi}(\vz | \vx)}\right],\nonumber\\
    &\pi_{\beta_{k}}(\vz) := p_{\theta}(\vx, \vz)^{\beta}q_{\phi}(\vz|\vx)^{1-\beta}/Z_{\beta}, \nonumber&&
    Z_{\beta} := \int p_{\theta}(\vx, \vz)^{\beta}q_{\phi}(\vz|\vx)^{1-\beta} \vdz.
\end{align}

\section{Maximizing the TVO minimizes a divergence between the variational distribution and true posterior}
\label{app:divergence}
We now show:
\begin{align}
    \TVO(\theta, \phi, \vx) = \log p_\theta(\vx) - \mathcal{D}(q_{\phi}(\vz|\vx) ||p_{\theta}(\vz | \vx))
\end{align}

Where $\mathcal{D}(q_{\phi}(\vz|\vx) ||p_{\theta}(\vz | \vx))$ is a divergence between the variational distribution $q_{\phi}(\vz|\vx)$ and true posterior $p_{\theta}(\vz | \vx)$. We refer to the notation defined in~\cref{app:integrand_notation} and the definition of divergence defined by~\citet{eguchi1985differential}.

\begin{proof}[Proof]
    The \gls{TVO} is a left Riemann sum approximation of $\log p_\theta(\vx) = \int_0^1 g(\beta) d\beta$, where $g(\beta) = \E_{\pi_\beta(\vz)}\left[ U^{\prime}(\vz) \right]$ and $g(\beta)$ is a differentiable monotonically non-decreasing function in $\beta$ (cf. Equation~\eqref{eq:gvar_equivalence}). The \gls{TVO} is therefore a lower bound of $\log p_\theta(\vx)$ and can be written
\begin{align}
    \TVO(\theta, \phi, \vx)&\le \log p_\theta(\vx) \nonumber\\
    \TVO(\theta, \phi, \vx)&= \log p_\theta(\vx) - c(\theta, \phi, \vx),~~c(\theta, \phi, \vx) \ge 0\label{app:divergence/tvo_constant}
\end{align}
We will show $c(\theta, \phi, \vx) = \mathcal{D}(q_{\phi}(\vz|\vx) ||p_{\theta}(\vz | \vx))$, which is equivalent to showing
\begin{enumerate}[label=\large\protect\textcircled{\small\arabic*}]
    \item $c \ge 0,~\forall~p_{\theta}(\vz | \vx), q_{\phi}(\vz|\vx) \in \mathcal{Z}$
    \item $c = 0 \iff p_{\theta}(\vz | \vx) = q_{\phi}(\vz|\vx)$
\end{enumerate}

\textcircled{\tiny{1}} is implied in the definition of $c$ in~\ref{app:divergence/tvo_constant}. We now show \textcircled{\tiny{2}}.

\paragraph{Forward direction} $\big(c = 0\big)  \Rightarrow \big(p_{\theta}(\vz | \vx) = q_{\phi}(\vz|\vx)\big)$

If $c = 0$, the left Riemann sum must be an exact approximation to $\int_0^1 g(\beta) d\beta$. Because is $g(\beta)$ is differentiable (and assuming it is finite), the Riemann approximation can only be exact when $g(\beta)$ is flat (i.e. $\frac{\partial g(\beta)}{\partial \beta} = 0)$ in the region $\beta \in [0, 1]$. We first recall that by definition, $\pi_0(\vz) = q_{\phi}(\vz|\vx)$ and $\pi_1(\vz) = p_{\theta}(\vz|\vx)$. Therefore
\begin{align}
    \int_0^1 \frac{\partial g(\beta)}{\partial \beta} d\beta &= \int_0^1 0~d\beta\\
    g(1) - g(0) &= 0\\
    g(1) &= g(0) \\
    \E_{\pi_1(\vz)}\left[ U^{\prime}(\vz) \right] &= \E_{\pi_0(\vz)}\left[ U^{\prime}(\vz) \right]\\
    \E_{p_{\theta}(\vz | \vx)}\left[ U^{\prime}(\vz) \right] &= \E_{q_{\phi}(\vz|\vx)}\left[ U^{\prime}(\vz) \right]
\end{align}
Which is only possible when $p_{\theta}(\vz | \vx) = q_{\phi}(\vz|\vx)$.

\paragraph{Reverse direction} $\big(p_{\theta}(\vz | \vx) = q_{\phi}(\vz|\vx)\big) \Rightarrow \big(c = 0\big)$

If $p_{\theta}(\vz | \vx) = q_{\phi}(\vz|\vx)$, the \gls{TVO} can be written as
\begin{align}
    \TVO(\theta, \phi, \vx) &= \frac{1}{K}\sum_{k=0}^{K-1}\E_{\pi_{\beta_{k}}(\vz)}\left[\log \frac{p_{\theta}(\vx, \vz)}{p_{\theta}(\vz | \vx)}\right] \\
     &= \frac{1}{K}\sum_{k=0}^{K-1}\E_{\pi_{\beta_{k}}(\vz)}\left[\log p_\theta(\vx)\right] \\
     &= \log p_\theta(\vx)
\end{align}
Therefore $c = 0$.
\end{proof}
\section{Derivation of the Covariance Gradient Estimator}
\label{app:gradient_estimator_derivation}

We want to show that
\begin{align}
    \nabla_\lambda \E_{\pi_{\lambda, \beta}}[f(\vz, \lambda)] = \E_{\pi_{\lambda, \beta}}[\nabla_\lambda f(\vz, \lambda)] + \Cov_{\pi_{\lambda, \beta}}\left[\nabla_\lambda \log \tilde{\pi}_{\lambda, \beta}(\vz), f(\vz, \lambda)\right]. \label{eq:thingwewanttoprove}
\end{align}

Our estimator holds under the common regularity conditions assumed for the score function estimator~\citet{l1993interchange}. We begin with a simple lemma.

  \begin{lemma}
    \begin{align}
      \nabla_{\lambda} \log Z_{\lambda, \beta}(\vx) &= \E_{\pi_\beta(\vz)}[\nabla_{\lambda} \log \tilde{\pi}_{\lambda, \beta}(\vz)]
    \end{align}
    \begin{proof}[Proof of lemma 1]
      \begin{align}
        \nabla_{\lambda} \log Z_{\lambda, \beta}(\vx) &= \textcolor{blue}{\frac{1}{Z_{\lambda, \beta}(\vx)} \nabla_{\lambda}} Z_{\lambda, \beta}(\vx) \\
        &= \frac{1}{Z_{\lambda, \beta}(\vx)} \nabla_{\lambda} \textcolor{blue}{\int \tilde{\pi}_{\lambda, \beta} (\vz)\vdz} \\
        &= \frac{1}{Z_{\lambda, \beta}(\vx)} \textcolor{blue}{\int \nabla_{\lambda}}\tilde{\pi}_{\lambda, \beta} (\vz)\vdz \\
        &= \textcolor{blue}{\int \frac{\tilde{\pi}_{\lambda, \beta}(\vz)}{Z_{\lambda, \beta}(\vx)} \nabla_{\lambda} \log} \tilde{\pi}_{\lambda, \beta}(\vz) \vdz \\
        &= \textcolor{blue}{\E_{\pi_\beta(\vz)}}[\nabla_{\lambda}\log \tilde{\pi}_{\lambda, \beta}(\vz)]
      \end{align}
    \end{proof}
  \end{lemma}

To prove \eqref{eq:thingwewanttoprove}, we use the product rule and rearrange
\begin{align}
    \nabla_{\lambda}\E_{\pi_\beta(\vz)}[f(\vz, \lambda)] &= \E_{\pi_\beta(\vz)}[\nabla_{\lambda} f(\vz, \lambda) + f(\vz, \lambda) \nabla_{\lambda} \log \pi_{\lambda, \beta}(\vz|\vx)]\\
    = \E_{\pi_\beta(\vz)}[\nabla_{\lambda} f(\vz, \lambda) &+ f(\vz, \lambda) \textcolor{blue}{\big(\nabla_{\lambda} \log \tilde{\pi}_{\lambda, \beta}(\vz) - \nabla_{\lambda} \log Z_{\lambda, \beta}(\vx)\big)} ] \\
    = \E_{\pi_\beta(\vz)}[\nabla_{\lambda} f(\vz, \lambda)] &+ \textcolor{blue} {\E_{\pi_\beta(\vz)}}[f(\vz, \lambda) \nabla_{\lambda} \log \tilde{\pi}_{\lambda, \beta}(\vz)] \nonumber \\
    &- \textcolor{blue}{\E_{\pi_\beta(\vz)}}[f(\vz, \lambda) \nabla_{\lambda} \log Z_{\lambda, \beta}(\vx)].
\end{align}

Now using lemma 1 on the third term
\begin{align}
    = \E_{\pi_\beta(\vz)}[\nabla_{\lambda} f(\vz, \lambda)] &+ \E_{\pi_\beta(\vz)}[f(\vz, \lambda) \nabla_{\lambda} \log \tilde{\pi}_{\lambda, \beta}(\vz)] \nonumber \\
    &- \E_{\pi_\beta(\vz)}[f(\vz, \lambda)]\textcolor{blue}{\E_{\pi_\beta(\vz)}[\nabla_{\lambda}\log \tilde{\pi}_{\lambda, \beta}(\vz)]}\\
    = \E_{\pi_\beta(\vz)}[\nabla_{\lambda} f(\vz, \lambda)] & + \textcolor{blue}{\Cov_{\pi_{\lambda, \beta}(\vz|\vx)}} \big[\nabla_{\lambda} \log \tilde{\pi}_{\lambda, \beta}(\vz), f(\vz, \lambda) \big].
\end{align}

\section{Variance of the Covariance Gradient Estimator and its Relationship to \textsc{reinforce}}
\label{app:covariance_estimator_properties}

In this section we clarify the difference between the covariance estimator \eqref{eq:method/estimation/grad} and the \textsc{reinforce} estimator and empirically investigate its variance.

While both estimators use the log-derivative trick, the main difference between the two is the \textsc{reinforce} estimator requires differentiating through $\log \pi_{\beta}(\vz)$ which contains the intractable normalizing constant, while the covariance estimator only requires differentiating through the unnormalized distribution $\log \tilde{\pi}_{\beta}(\vz)$.

We can state the difference as follows. Assuming $\pi_\beta(\vz) = \tilde{\pi}_\beta(\vz)/Z_\beta$ depends on parameters $\lambda$, to compute $\nabla_{\lambda} \E_{\pi_\beta(\vz)}\left[f(\vz)\right]$, one can use the following gradient estimators:
\begin{flalign*}
    \textsc{reinforce:} &&&\E_{\pi_\beta(\vz)}\left[f(\vz) \nabla_{\lambda}\log \pi_\beta(\vz) \right]&& & \\
    \textsc{reinforce + baseline:} &&&\E_{\pi_\beta(\vz)}\left[\left(f(\vz) - \textcolor{blue}{\E_{\pi_\beta(\vz)}\left[ f(\vz) \right]} \right) \nabla_{\lambda}\log \pi_\beta(\vz) \right]&& &\\
    \textsc{cov. estimator} \text{ (ours):} &&&\E_{\pi_\beta(\vz)}\left[\left(f(\vz) - \E_{\pi_\beta(\vz)}\left[ f(\vz) \right] \right) \left(\nabla_{\lambda}\log \textcolor{blue}{\tilde{\pi}_\beta(\vz)} - \textcolor{blue}{\E_{\pi_\beta(\vz)}\left[\nabla_{\lambda}\log \tilde{\pi}_\beta(\vz) \right] }  \right) \right]&& &
\end{flalign*}

Unlike \textsc{reinforce}, where a baseline is typically added ad-hoc to reduce variance, the baseline $b = \E_{\pi_\beta(\vz)}\left[ f(\vz) \right]$ naturally appears as a result of differentiating through $\pi_\beta(\vz)$ using the identity $\nabla_{\lambda} \log Z_{\lambda, \beta}(\vx) = \E_{\pi_\beta(\vz)}[\nabla_{\lambda} \log \tilde{\pi}_{\lambda, \beta}(\vz)]$ derived in appendix E.
The baseline also partially explains the low variance of our estimator, as it is equivalent to the ``average baseline'' often used reinforcement learning~\citep{greensmith2004variance, weaver2001optimal}.

A second source of variance-reduction comes from reusing samples, a method known as ``common random numbers''~\citep{owen2013monte}.
The terms in the \gls{TVO} are highly correlated, thus we expect reusing a single batch of samples for each additional term will act to reduce variance according to equation 8.21 in~\citet{owen2013monte}.
However, because the covariance term breaks into both positive and negative terms, common random numbers could potentially increase variance. In \cref{tab:common_random_numbers} we show the average gradient std at different iterations during the training procedure, using the $S=50$ discrete model described in \cref{sec:discrete_experiment} and in \cref{fig:experiments/discrete_vae/baselines} and \cref{fig:experiments/discrete_vae/kl_and_grad_std}. It is evident reusing samples significantly reduces the variance of the covariance gradient estimator, often by more than a factor of two.

\renewcommand\arraystretch{1.5}

\begin{table}[]
    \centering % used for centering table
    \caption{The effect of Common Random Numbers (CRN) on TVO variance. We use the discrete model of \cref{sec:discrete_experiment}}
    \label{tab:common_random_numbers}
    \begin{tabular}{llllll}
        \toprule
        Iterations           & 10    & 1e6  & 2e6  & 3e6  & 4e6  \\ \midrule
        Gradient std w/o CRN & 52.33 & 2.88 & 2.57 & 2.39 & 2.47 \\
        Gradient std w/ CRN  & 8.19  & 1.38 & 1.17 & 1.05 & 1.03 \\ \bottomrule
    \end{tabular}
\end{table}

\begin{figure}[t]
    \centering
    \includegraphics[width=0.80\textwidth]{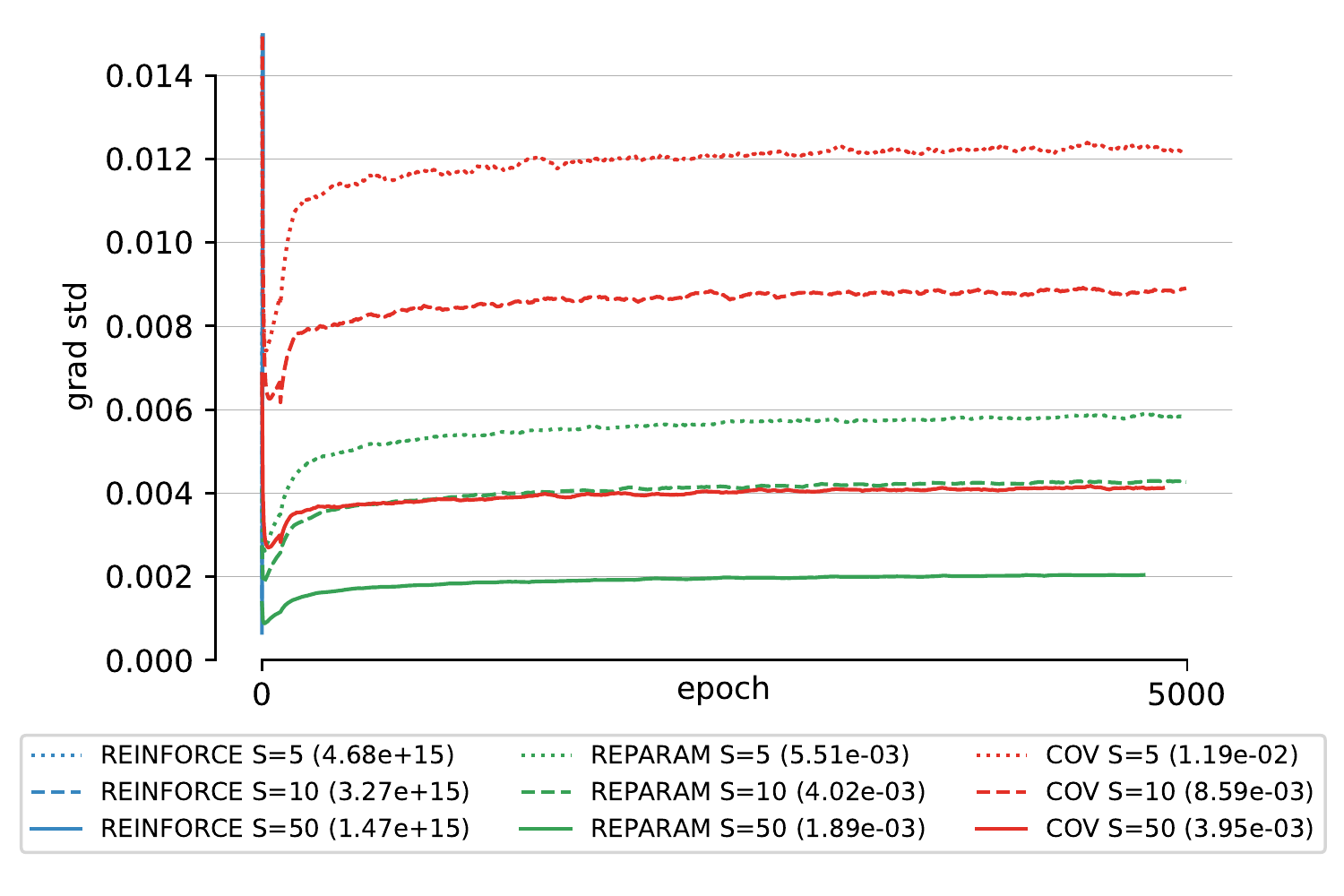}
    \caption{Comparing the standard deviation of gradient estimators on continuous \glspl{VAE} trained on the \gls{ELBO}. The covariance estimator has higher variance than the reparameterization trick for all $S$ but much lower than \textsc{reinforce}, which is numerically unstable.}
    \label{fig:appendix/elbo_grad_variance}
\end{figure}

In \cref{fig:appendix/elbo_grad_variance} we compare the variance of our estimator to the reparameterization trick and \textsc{reinforce} on the continuous model described in \cref{sec:continuous_experiment}. To control for any possible effect on variance the additional terms in the \gls{TVO} could have, we use the $\ELBO$ (i.e the \gls{TVO} with $K=1$), and plot the gradient standard deviation for the \textsc{cov} estimator (ours), the reparameterization trick and \textsc{reinforce}. The \textsc{cov} estimator has higher variance than the reparameterization trick estimator, and outperforms the \textsc{reinforce} estimator which is numerically unstable. Both the standard deviation of both the \textsc{cov} estimator and the reparameterization trick improves as samples increase but the effect is more prominent for the \textsc{cov} estimator.

\section{Special Cases of the TVO}
\label{app:special_cases}

In \cref{tab:generalizations}, we summarize the different ways the \gls{TVO} generalizes existing variational objectives, and in the following subsection we list the mathematical form of each objective.
In the main text, we mentioned that the lower bound variant of the \gls{TVO} with $K = 1$ partition can be seen as the \gls{ELBO}. This connects the \gls{TVO} to all methods that maximize the \gls{ELBO}. The upper bound variant of the \gls{TVO} with $K = 1$ partition can be seen as \gls{EUBO}. This therefore connects the \gls{TVO} to all methods that minimize the reverse KL divergence $\KL{p_\theta(\vz \given \vx)}{q_\phi(\vz \given \vx)}$, including \gls{WS}, \gls{RWS} and inference compilation.

For $K > 1$, we have a novel objective which we can optimize with respect to $\theta$, $\phi$ or both and therefore apply to all the variational methods summarized in \cref{tab:generalizations}.

\begin{table}[ht]
    \centering
    \caption{The thermodynamic variational identity generalizes existing variational objectives.}
    \label{tab:generalizations}
    \begin{tabular}{@{}cccccc@{}}
    \toprule
    \multicolumn{2}{c}{Approximation} & \multicolumn{2}{c}{\makecell{Left Riemann sum\\(lower bound---maximize)}} & \multicolumn{2}{c}{\makecell{Right Riemann sum\\(upper bound---minimize)}} \\ \cmidrule(l){3-4}\cmidrule(l){5-6}
    \multicolumn{2}{c}{Number of partitions} & $1$ & $>1$ & $1$ & $>1$ \\ \midrule
    \multicolumn{1}{c}{\multirow{5}{*}{\vspace{-2.6ex}Optimize}}
    & $\theta$ & wake in \textsc{ws} & $\TVO_K^L(\theta, \vx)$ & N/A & N/A \\ \cmidrule(l){2-6}
    & $\phi$ & \textsc{vi} & $\TVO_K^L(\phi, \vx)$ & \makecell{wake-$\phi$ in \textsc{rws}, \\ sleep in \textsc{ws}, \\ inference compilation} & $\TVO_K^U(\theta, \phi, \vx) $ \\ \cmidrule(l){2-6}
    & $\theta, \phi$ & \textsc{vae} & $\TVO_K^L(\theta, \phi, \vx)$ & N/A & N/A \\ \bottomrule
\end{tabular}
\end{table}

\subsection{Variational Objective Zoo}

In the following we show how a number of commonly used variational objectives can be recovered from the \gls{TVO} using a single partition $K=1$. Each method can be extended by setting $K > 1$.

We have three degrees of freedom: 1) Whether we optimize $\theta, \phi$, or both 2) whether we maximize $\TVO_1^L(\theta, \phi, \vx)$ or minimize $\TVO_1^U(\theta, \phi, \vx)$, and 3) whether we use data sampled from the true data distribution $\{\vx_i\}_{i=1}^{N} \overset{\text{i.i.d}}{\sim} p(\vx)$ or from our generative model $\{\vx_i\}_{i=1}^{N} \overset{\text{i.i.d}}{\sim} p_\theta(\vx)$, as in the case of inference compilation and the sleep phase of the wake-sleep algorithm.
\paragraph{Variational Inference}
Variational inference~\citep{blei2017variational} can be recovered by learning $\phi$ and maximizing $\TVO_1^L(\phi, \vx)$ using real data:
\begin{align}
    \phi^* &= \argmax_{\phi} \left[\TVO_1^L(\phi, \vx)\right] \\
           &= \argmax_{\phi} \left[(1 - 0)\E_{\pi_0} \left[\log \frac{p(\vx,\vz)}{q_{\phi}(\vz | \vx)}\right] \right] \\
           &= \argmax_{\phi} \ELBO(\phi, \vx)
  \end{align}
\paragraph{Inference Compilation}
If we instead sample data from our generative model $\{\vx_i\}_{i=1}^{N} \sim p_{\theta}(x)$ and minimize $\TVO_1^U(\phi, \vx)$ we recover the inference compilation objective~\citep{le2016inference}:
\begin{align}
    \phi^* &= \argmin_{\phi} \E_{x \sim p_{\theta}(\vx)}\left[\TVO_1^U(\phi, \vx)\right]\\
    &=\argmin_{\phi} \int p_{\theta}(\vx) \left[\E_{p_{\theta}(\vz|\vx)} \left[\log \frac{p(\vx,\vz)}{q_{\phi}(\vz | \vx)}\right]\right] \vdx\\
    &=\argmin_{\phi} \int p_{\theta}(\vx) \left[\int \frac{p_{\theta}(\vx, \vz)}{p_{\theta}(\vx)} \left[\log \frac{p(\vx,\vz)}{q_{\phi}(\vz | \vx)}\right]\right]\vdz \vdx \\
    &=\argmin_{\phi} \int \int p_{\theta}(\vx, \vz) \left[\log \frac{p(\vx,\vz)}{q_{\phi}(\vz | \vx)}\right]\vdz \vdx \\
    &=\argmin_{\phi} \E_{p(\vx, \vz)}\left[-\log q_{\phi}(\vz | \vx)\right]
\end{align}
\paragraph{Variational Autoencoders}
The loss for \glspl{VAE}\citep{kingma2014auto,rezende2014stochastic} follows the same setting as in the variational inference objective, above except now we learn both $\phi$ and $\theta$ and average over data sampled from the true data distribution $\{\vx_i\}_{i=1}^{N} \overset{\text{i.i.d}}{\sim} p(\vx)$.
\begin{align}
  \phi^* &= \argmax_{\phi, \theta} \E_{x \sim p(\vx)}\left[\TVO_1^L(\theta, \phi, \vx)\right] \\
         &= \argmax_{\phi, \theta} \frac{1}{N}\sum_{i=1}^N\ELBO(\theta, \phi, \vx_i)
\end{align}

\paragraph{Wake-sleep and reweighted wake sleep}

In the original wake-sleep algorithm~\citep{hinton1995wake}, the authors proposed the \textit{wake-phase $\theta$} update and \textit{sleep-phase $\phi$} updates to train the generative model and inference network respectively. In Reweighted Wake-Sleep~\citep{bornschein2015reweighted}, two more objectives were proposed, the \textit{reweighted wake-phase $\theta$} update\footnote{This was not the authors' original terminology and is used here to differentiate this objective from the original wake-phase $\theta$ update.} and the \textit{wake-phase $\phi$} update. All except the \textit{reweighted wake-phase $\theta$}\footnote{This objective is not a special case of the \gls{TVO} and is therefore not included in \cref{tab:generalizations}} are special cases of the \gls{TVO} and are listed below.

\begin{itemize}
    \item \textbf{Wake-phase $\theta$ update}
In the wake phase $\theta$ update, we consider $\phi$ fixed and maximize $\TVO_1^L(\theta, \vx)$, using data $\{\vx_i\}_{i=1}^{N} \overset{\text{i.i.d}}{\sim} p(\vx)$ sampled from the true distribution. This is similar to the variational inference update except we're learning $\theta$ instead of $\phi$:
\begin{align}
  \theta^* &= \argmax_{\theta} \E_{x \sim p(\vx)}[\TVO_1^L(\theta, \vx)] \\
           &= \argmax_{\theta} \frac{1}{N}\sum_{i=1}^N\ELBO(\theta, \vx_i)
\end{align}
\item \textbf{Sleep-phase $\phi$ update}
In the sleep phase $\phi$ update, we consider $\theta$ fixed and minimize $\TVO_1^U(\phi, \vx)$ using simulated data $\{\vx_i\}_{i=1}^{N} \sim p_{\theta}(\vx)$ and a single partition. This objective is the same as the inference compilation objective.
\item \textbf{Wake-phase $\phi$ update}
In the wake phase $\phi$ update, we instead use real data $\{\vx_i\}_{i=1}^{N} \sim p(\vx)$ and again minimize $\TVO_1^U$:
\begin{align}
  \phi^* &= \argmin_{\phi} \E_{\vx \sim p(\vx)}\left[\TVO_1^U(\phi, \vx)\right]\\
  &= \argmin_{\phi} \E_{\vx \sim p(\vx)}\left[\E_{p_{\theta}(\vz|\vx)}\left[\log \frac{p_{\theta}(\vx,\vz)}{q_{\phi}(\vz | \vx)}\right]\right]\\
  &= \argmin_{\phi} \E_{\vx \sim p(\vx)}\left[\E_{p_{\theta}(\vz|\vx)}\left[-\log q_{\phi}(\vz | \vx)\right]\right]
\end{align}

This is the objective given in the wake-phase $\phi$ update in equation 6 of \citet{le2018revisiting}. The gradient estimator for performing this update given in~\citet{le2018revisiting} is equivalent to the gradient estimator obtained via equations \eqref{eq:method/estimation/grad} and \eqref{eq:method/estimation/general_functions}.
\end{itemize}

\section{Additional Illustrations of the Thermodynamic Variational Identity}
\label{app:additional_illustrations}

\begin{figure}[!ht]
    \begin{minipage}{0.33\textwidth}
      \centering
      \includegraphics[width=\textwidth]{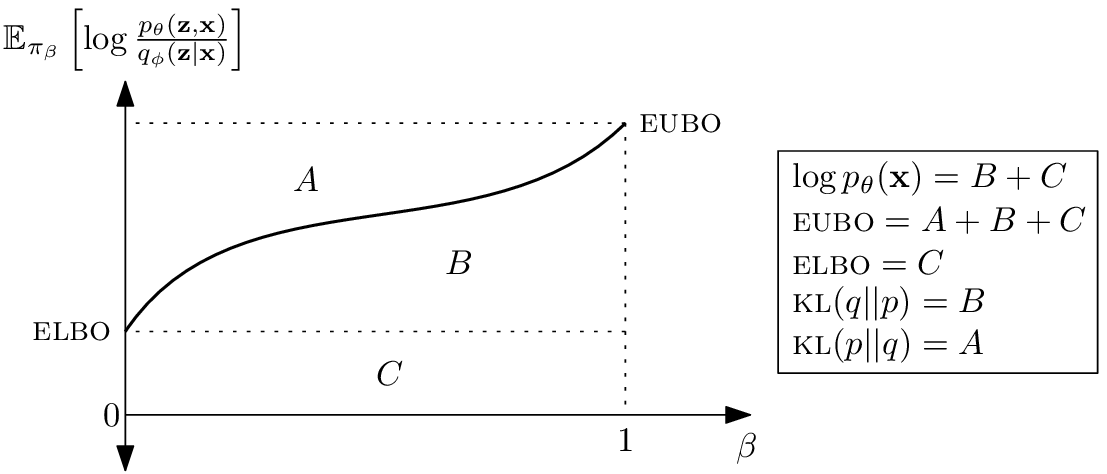}
    \end{minipage}\hfill
    \begin{minipage}{0.33\textwidth}
      \centering
      \includegraphics[width=\textwidth]{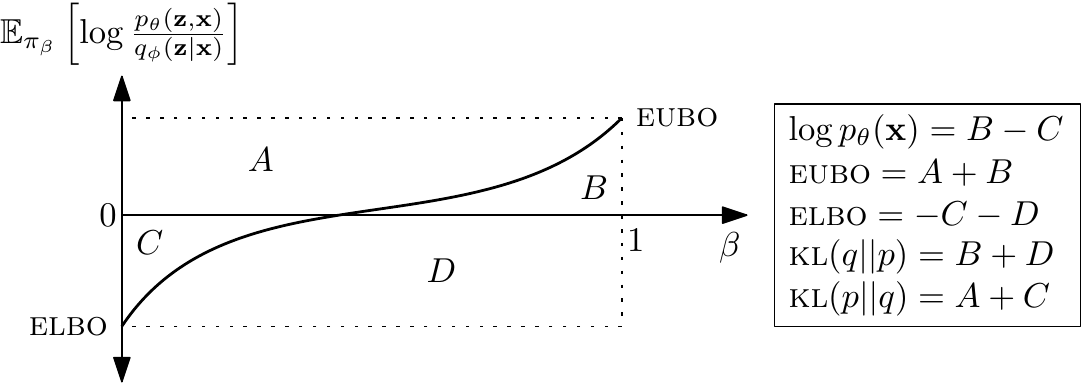}
    \end{minipage}\hfill
    \begin{minipage}{0.33\textwidth}
      \centering
      \includegraphics[width=\textwidth]{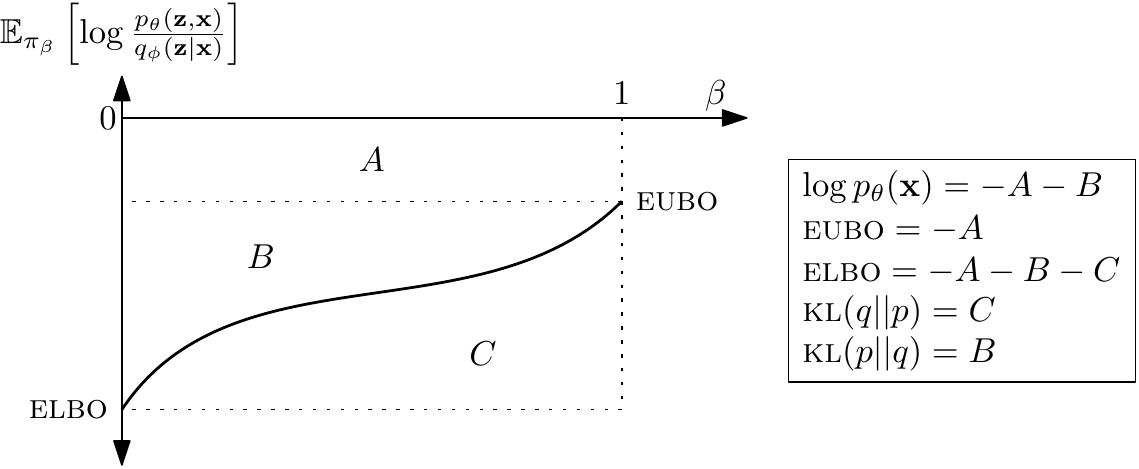}
    \end{minipage}
    \caption{Different scenarios of the $\E_{\pi_\beta}[U^{\prime}(\vz)]$ curve where $\ELBO < 0$. On the left, $0 < \ELBO < \EUBO$. In the middle, $\ELBO < 0 < \EUBO$. On the right $\ELBO < \EUBO < 0$.}
    \label{fig:additional_illustrations/thermo_diagrams_extra}
 \end{figure}

In \cref{fig:additional_illustrations/thermo_diagrams_extra}, we provide illustrations of how the $\E_{\pi_\beta}[U^{\prime}(\vz)]$ curve relates to $\log p_\theta(\vx)$, $\KL{q}{p}$, $\KL{p}{q}$, \gls{ELBO} and \gls{EUBO} for the cases of $\ELBO < 0 < \EUBO$ and $\ELBO < \EUBO < 0$.
In the following, we provide derivations to justify the illustrations.

\paragraph{Case $\ELBO < 0 < \EUBO$}
The top-most point of the curve is the $\EUBO$ by definition which means that the area $A + B$ is equal to the $\EUBO$ because of the unit length of the rectangle.
In a similar manner, the $\ELBO$ is the negative of the area of $C + D$.
Now, due to the thermodynamic identity,  $\log p_\theta(\vx) = \int_{\beta = 0}^1 \E_{\pi_\beta}[U^{\prime}(\vz)] \,\mathrm d\beta$, it is equal to $B - C$ which is the area denoted by the definite integral.

To obtain the expressions for the \gls{KL}, we use the identities
\begin{align}
\log p_\theta(\vx) &= \ELBO(\vx, \theta, \phi) + \KL{q_\phi(\vz \given \vx)}{p_\theta(\vz \given \vx)} \\
&= \EUBO(\vx, \theta, \phi) - \KL{p_\theta(\vz \given \vx)}{q_\phi(\vz \given \vx)}
\end{align}

\paragraph{Case $\ELBO < \EUBO < 0$}
The top-most point of the curve is the $\EUBO$ by definition which means that $-A$ is equal to the $\EUBO$ because of the unit length of the rectangle.
In a similar manner, the $\ELBO$ is $-A - B - C$.
Due to the thermodynamic identity,  $\log p_\theta(\vx) = \int_{\beta = 0}^1 \E_{\pi_\beta}[U^{\prime}(\vz)] \,\mathrm d\beta$, it is equal to $-A - B$ which is the area denoted by the definite integral.
We obtain expressions for the \gls{KL} similarly as before.

Similar line of reasoning gives rise to the relationships in \cref{fig:additional_illustrations/thermo_diagrams_extra} (left).

\section{Details for Deep Generative Models}
\label{app:vae_details}

\paragraph{Discrete latent variables.}
Sigmoid belief networks are used to evaluate objectives, continuous relaxations and control variate methods for learning discrete latent variable models~\citep{mnih2014neural,maddison2017concrete,jang2017categorical,mnih2016variational,bornschein2015reweighted,tucker2017rebar,grathwohl2018backpropagation}.
The generative model is of the form $p(\vz_{1:L}, \vx) = p(\vz_L) \prod_{\ell = 1}^{L - 1} p(\vz_\ell \given \vz_{\ell + 1}) p(\vx \given \vz_1)$ where each conditional on $\vz_\ell$ is an independent Bernoulli whose parameters are a linear function of $\vz_{\ell + 1}$.
The likelihood $p(\vx \given \vz_1)$ is also an independent Bernoulli whose parameters are a linear function of $\vz_1$ and we parameterize the prior $p(\vz_L)$.
\begin{align*}
    p_\theta(\vz_L) &= \mathrm{Bernoulli}(\vz_L \given \mathbf b_L), \\
    p_\theta(\vz_\ell \given \vz_{\ell + 1}) &= \mathrm{Bernoulli}(\vz_\ell \given \mathrm{decoder}_\ell(2\vz_{\ell + 1} - 1)) & \ell = L - 1, \dotsc, 1, \\
    p_\theta(\vx \given \vz_1) &= \mathrm{Bernoulli}(\vx \given \mathrm{decoder}_x(2 \vz_1 - 1) + \tilde{\vx})
\end{align*}
The inference network is factorized in the opposite way to the generative model, where $q(\vz \given \vx) = q(\vz_1 \given \vx) \prod_{\ell = 2}^L q(\vz_\ell \given \vz_{\ell - 1})$.
Here, each conditional is an independent Bernoulli whose parameters are linear functions of the condition.
\begin{align*}
    q_\phi(\vz_1 \given \vx) &= \mathrm{Bernoulli}\left(\vz_1 \bigg\given \mathrm{encoder}_1\left(\frac{\vx - \bar{\vx} + 1}{2}\right)\right), \\
    q_\phi(\vz_\ell \given \vz_{\ell - 1}) &= \mathrm{Bernoulli}(\vz_\ell \given \mathrm{encoder}_\ell(2\vz_{\ell - 1} - 1)) & \ell = 2, \dotsc, L,
\end{align*}
where $\vx \in \{0, 1\}^{D_x}$ and $\vz_\ell \in \{0, 1\}^{D_z}$. We set $L = 2$, $D_x = 784$ and $D_z = 200$.
We used Pytorch's default parameter initialization.
The Bernoulli distributions are independent Bernoulli distributions whose parameters are logits, i.e. they get passed through a sigmoid function to obtain the probability.
$\bar{\vx}$ is the mean over training data set and $\tilde{\vx} = \log\left(\bar{\vx} - 1\right)$.
In the linear case, the encoders and decoders are linear functions of their inputs.
In the non-linear case, they are a three-layer multilayer perceptrons with $\tanh$ nonlinearities of the form $\texttt{input\_dim} \xrightarrow{\mathrm{Lin} + \tanh} D_z \xrightarrow{\mathrm{Lin} + \tanh} D_z \xrightarrow{\mathrm{Lin}} \texttt{output\_dim}$.

We used the Adam optimizer with the learning rate in $\{3 \times 10^{-4}, 10^{-3}, 3 \times 10^{-3}\}$ and the other hyperparameters being set to the defaults.
We picked the learning rate which performed best on the validation set which was $3 \times 10^{-4}$ for all algorithms.
We ran the optimization for $4$ million iterations with batch size $24$.

\paragraph{Continuous latent variables.}
The model is of the form $p(\vz)p_\theta(\vx \given \vz) = \mathrm{Normal}(\vz \given 0, I) \mathrm{Bernoulli}(\vx \given \mathrm{decoder}_\theta(\vz))$, where $\vz$ is $200$-dimensional and $\mathrm{decoder}_\theta$ is a three-layer multilayer perceptron with $\tanh$ activations and sigmoid output which parameterizes the probabilities of the independent Bernoulli distribution.
\begin{align*}
    p(\vz) &= \mathrm{Normal}(\vz \given 0, I), \\
    p_\theta(\vx \given \vz) &= \mathrm{Bernoulli}(\vx \given \mathrm{decoder}_\theta(\vz))
  \end{align*}
The inference network is of the form $q_\phi(\vz \given \vx) = \mathrm{Normal}(\vz \given \mathrm{encoder}_\phi(\vx))$, where the encoder is a two-layer multilayer perceptron with $\tanh$ activations. The output is passed through two separate linear layers which output the mean and the log standard deviations of the independent normal distribution.
\begin{align*}
  q_\phi(\vz \given \vx) &= \mathrm{Normal}(\vz \given \mathrm{encoder}_\phi(\vx)),
\end{align*}
where $\vx \in \{0, 1\}^{D_x}$ and $\vz \in \mathbb R^{D_z}$ for $D_x = 784$ and $D_z = 200$.
The decoder is of the form $D_z \xrightarrow{\mathrm{Lin} + \tanh} D_z \xrightarrow{\mathrm{Lin} + \tanh} D_z \xrightarrow{\mathrm{Lin}} D_x$ and its output is passed through a sigmoid to obtain probabilities for the Bernoulli distribution.
The encoder is of the form $D_x \xrightarrow{\mathrm{Lin} + \tanh} D_z \xrightarrow{\mathrm{Lin} + \tanh} D_z$.
Its output is passed through two \emph{separate} neural networks of the form $D_z \xrightarrow{\mathrm{Lin}} D_z$ which output the means and log standard deviations of the independent Normal distribution.
\newpage
\section{Notation}
\begin{table}[h!]\caption{Table of Notation}
  \begin{center}% used the environment to augment the vertical space
  % between the caption and the table
  \begin{tabular}{r c p{7cm}}
  \toprule
  $\{\vx_i\}_{i=1}^{N} $ & $:=$ & Data set consisting of N i.i.d samples $\vx_i \in \mathbb{R}^D$\\
  \addlinespace[0.1cm]
  $\{\vz_i\}_{i=1}^{N}$ & $:=$ & Unobserved latent random variables $\vz_i \in \mathbb{R}^M$\\
  \addlinespace[0.1cm]
  $p_{\theta}(\vx, \vz) = p_{\theta}(\vx|\vz)p_{\theta}(\vz)$ & $:=$ & The joint model parameterized by $\theta$, which factorizes into a likelihood $p_{\theta}(\vx|\vz)$ and prior $p_{\theta}(\vz)$\\
  % $p(\vz | \vx) = p_{\theta}(\vx, \vz) / Z_{\theta}(\vx)$ & $:=$ & The true (often intractable) posterior\\
  $p_{\theta}(\vz | \vx) = p_{\theta}(\vx, \vz) / p_{\theta}(\vx)$ & $:=$ & The true (often intractable) posterior\\
  $q_{\phi}(\vz|\vx)$ & $:=$ & The variational distribution parameterized by $\phi$. By assumption $q_{\phi}(\vz|\vx)$ is correctly normalized.\\
  \addlinespace[0.1cm]
  $\tilde{\pi}_{\lambda, \beta}(\vz) = p_{\theta}(\vx, \vz)^{\beta} q_{\phi}(\vz|\vx)^{1 - \beta}$ & $:=$ & The unnormalized path distributions. By construction, $\tilde{\pi}_{\lambda, \beta=1}(\vz) = p_{\theta}(\vx, \vz)$ and $\tilde{\pi}_{\lambda, \beta=0}(\vz|\vx) = q_{\phi}(\vz | \vx)$\\
  $\pi_{\lambda, \beta}(\vz|\vx) = \tilde{\pi}_{\lambda, \beta}(\vz) / Z_{\lambda, \beta}(\vx)$ & $:=$ & The path distributions parameterized by $\lambda~=~\{~\theta,~\phi~\}$ and scalar parameter $\beta \in [0, 1]$. By construction, $\pi_{\lambda, \beta=1}(\vz|\vx) = p_{\theta}(\vz | \vx)$ and $\pi_{\lambda, \beta=0}(\vz|\vx) = q_{\phi}(\vz | \vx)$\\
  $ Z_{\lambda, \beta}(\vx)= \int \tilde{\pi}_{\lambda, \beta}(\vz) \vdz_{1:N}$ & $:=$ & The normalizing constant for the path distributions. By construction $Z_{\lambda, \beta = 1}(\vx) = p_{\theta}(\vx)$ and $Z_{\lambda, \beta=0}(\vx) = 1$ (because $q_{\phi}(\vz|\vx)$ is assumed to be correctly normalized).\\
  $U_{\lambda, \beta}(\vz) = \log \tilde{\pi}_{\lambda, \beta}(\vz)$ & $:=$ & The potential energy.\\
  \addlinespace[0.1cm]
  $U^{\prime}_{\lambda, \beta}(\vz) = \frac{\partial}{\partial \beta} U_{\lambda, \beta}(\vz)$ & $:=$ & The first derivative of the potential w.r.t $\beta$, the inverse temperature. \\
  \bottomrule
  \end{tabular}
  \end{center}
  \label{tab:TableOfNotationForMyResearch}
  \end{table}

\section{Acronyms}
\textbf{AIS} Annealed Importance Sampling

\textbf{ELBO} Evidence Lower Bound

\textbf{EUBO} Evidence Upper Bound

\textbf{IS} Importance Sampling

\textbf{IWAE} Importance Weighted Autoencoder

\textbf{KL} Kullback Leibler

\textbf{RWS} Reweighted Wake Sleep

\textbf{SGD} Stochastic Gradient Descent

\textbf{TI} Thermodynamic Integration

\textbf{TVI} Thermodynamic Variational Identity

\textbf{TVO} Thermodynamic Variational Objective

\textbf{VAE} Variational Autoencoder

\textbf{VI} Variational Inference

\textbf{VIMCO} Variational Inference For Monte Carlo Objectives

\textbf{WS} Wake Sleep

% \acsetup{extra-style=comma}
% \printnoidxglossary[type=acronym]
% \printacronyms[title={test}]
\end{document}